\documentclass[]{article}
\usepackage{proceed2e}

\usepackage[latin9]{inputenc}
\usepackage{times}
\usepackage{array}
\usepackage{units}
\usepackage{multirow}
\usepackage{amssymb, amsmath, graphicx}
\usepackage{esint}
\usepackage[noadjust]{cite}
\usepackage{color}

\newcommand{\eqnref}[1]{Eq.~(\ref{#1})}

\newcommand{\lessdenselist}{
  \itemsep -1pt
}



\begin{document}

\title{Scalable Linear Causal Inference for Irregularly Sampled Time Series with Long Range Dependencies}

%\author{Francois W. Belletti, Evan R. Sparks, Joseph E. Gonzalez, Alexandre M. Bayen, Michael J. Franklin}

%ORIGINAL SUBMISSION.
% UAI  reviewing is double-blind.

% The author names and affiliations should appear only in the accepted paper.
%
\author{ 
{\bf Francois W. Belletti} \\
Computer Science Dept. \\
UC Berkeley\\
belletti@berkeley.edu\\
\And
{\bf Evan R. Sparks}  \\
Computer Science Dept. \\
UC Berkeley\\
sparks@cs.berkeley.edu \\
\And
{\bf Michael J. Franklin}   \\
Computer Science Dept.\\
UC Berkeley\\
franklin@berkeley.edu \\
\And
{\bf Alexandre M. Bayen}   \\
EECS Dept.\\
Inst. of Transportation Studies\\
UC Berkeley \\
bayen@berkeley.edu \\
\And
{\bf Joseph E. Gonzalez}   \\
Computer Science Dept.\\
UC Berkeley\\
jegonzal@berkeley.edu \\
}

\maketitle
% \begin{abstract}
% Analyzing the relation between correlated and lagged continuous
% stochastic processes is challenging as many classical second order 
% features such as cross-correlograms \cite{montgomery2015introduction, box2015time} 
% were defined in a setting in which all observations made simultaneously.
% With many large scale experimental data sets, \emph{Long Range Dependency} (LRD) \cite{doukhan2003theory} 
% additionally makes parallel processing in time domain impossible as the whole past 
% of the process needs to be taken into account when studying it at any time.
% Computing correlations directly on Brownian motions and fractional Brownian motions can be misleading for 
% practitioners \cite{phillips1986understanding} as the estimators have significant non vanishing variance.
% The new frequency domain based estimation work flow we propose is
% communication avoiding, uses limited memory and naturally handles irregularly sampled
% data without interpolation and computes fractional differences without full data shuffling. 
% In particular no sorting or re-organization of the data is
% necessary and no interpolation is used in time domain to fill gaps in observations. 
% Our method approximatively erases LRD with only linear complexity 
% with respect to the number of samples. 
% The efficiency of this new work flow is confirmed with Monte Carlo simulations and the
% study of high frequency trading financial records with Apache
% Spark \cite{zaharia2012resilient} as an execution engine in a distributed environment.
% \end{abstract}

\begin{abstract}
Linear causal analysis is central to a wide range of important application spanning finance, the physical sciences, and engineering.
Much of the existing literature in linear causal analysis operates in the \emph{time domain}.
Unfortunately, the direct application of time domain linear causal analysis to many real-world time series presents three critical challenges: \emph{irregular temporal sampling}, \emph{long range dependencies}, and \emph{scale}. 
Moreover, real-world data is often collected at irregular time intervals across vast arrays of decentralized sensors and with long range dependencies \cite{doukhan2003theory} which make naive time domain correlation estimators spurious \cite{granger1988causality}. 
% Analyzing the relation between correlated and lagged continuous
% stochastic processes is challenging as many classical second order 
% features such as cross-correlograms \cite{montgomery2015introduction, box2015time} 
% were defined in a setting in which all observations made simultaneously.
% With many large scale experimental data sets, \emph{Long Range Dependency} (LRD) \cite{doukhan2003theory} 
% additionally makes parallel processing in time domain impossible as the whole past 
% of the process needs to be taken into account when studying it at any time.
% Computing correlations directly on Brownian motions and fractional Brownian motions can be misleading for 
% practitioners \cite{phillips1986understanding} as the estimators have significant non vanishing variance.
In this paper we present a \emph{frequency domain} based estimation framework which naturally handles irregularly sampled data and long range dependencies while enabled memory and communication efficient distributed processing of time series data.
By operating in the frequency domain we eliminate the need to interpolate and help mitigate the effects of long range dependencies. 
We implement and evaluate our new work-flow in the distributed setting using Apache Spark and demonstrate on both Monte Carlo simulations and high-frequency financial trading that we can accurately recover causal structure at scale. % in large irregularly sampled time series. 
\end{abstract}

\vspace{-0.25cm}

\section{Introduction}

The analysis of time series is central to applications ranging from statistical finance \cite{abergel2012market,tsay2005analysis} to climate studies 
\cite{mudelsee2013climate} or cyberphysical systems such as the transportation network \cite{shang2005chaotic}.
In many of these applications one is interested in estimating 
the mutual linear predictive properties of events from time series data 
corresponding to a collection of data streams each of which is a series of pairs \textit{(timestamp, observation)}. 

\begin{figure}[h]
    \centering
        \includegraphics[width=.47\textwidth]{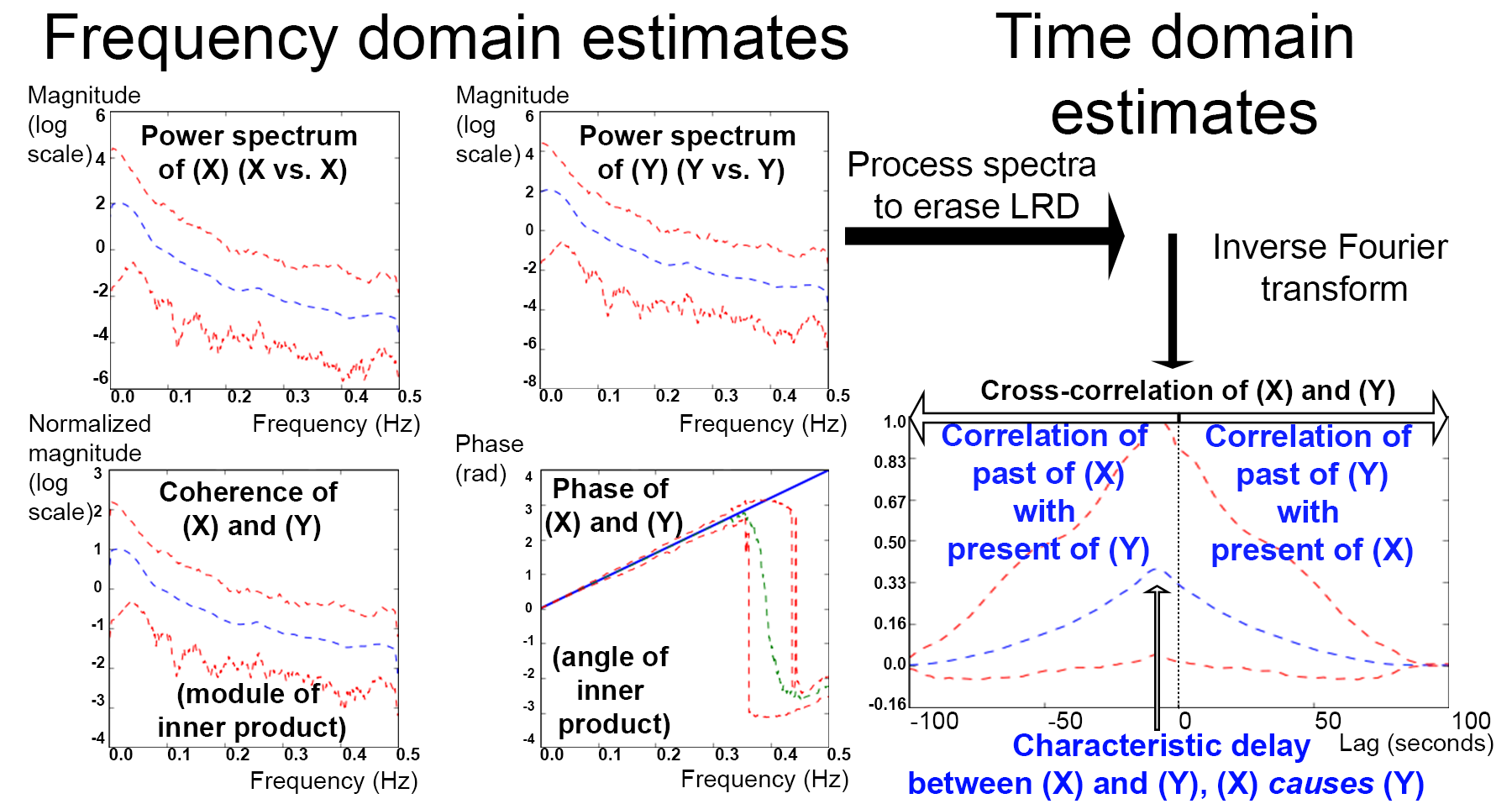}
    \caption{\footnotesize \textbf{Frequency domain causal analysis work flow} for two irregularly
sampled correlated and lagged Brownian motion increments and the derived cross-correlogram of incremements which highlights their causal structure. Dotted lines represent 5th, 50th and 95th percentiles respectively. Frequency domain estimation is treated in Section 2 and LRD erasure in Section 3. Once the cross-correlogram has been estimated, practitioners read out the direction of linear causality in the asymmetry of the curve. The lag value for which cross-correlation reaches a maximum can be interpreted as the characteristic delay of causation of the two processes.
}
    \label{fig:Illustration-cross-corre}
    \vspace{-0.3cm}
\end{figure}

%\par{\textbf{Discrete time processes:}}
%In \cite{granger1969investigating}, Granger introduced a statistical definition of linear causation as the assessment of the significance of a parametric model. Consider two centered real valued discrete time processes $(U_t)_{t \in \mathbb{Z}}$ and $(V_t)_{t \in \mathbb{Z}}$ which we assume are second order stationary \cite{harvey1993time}.
%The process $(U)$ is said to Granger-cause $(V)$ if in the linear regression
%\begin{equation}
%    V_t = \sum_{h = 0}^H a_h^{VV} V_{t-h} + \sum_{h = 0}^H a_h^{UV} U_{t-h}
%\end{equation}
%has some coefficients $a_h^{UV}$ statistically different from $0$.
%Such coefficients can classically be estimated by estimating the cross-correlogram of $U$ and $V$ for a finite number of lags and then solve the corresponding Yule-Walker equations \cite{Brockwell:1986:TST:17326}.
%Here we interpret causality to mean that past and present observations of a process $(X)$ \emph{predict} the future of another process $(Y)$. 
%We further restrict our attention to linear causal dependencies between continuous time stochastic processes which are both generally practical to infer and 
%provide a rich understanding of the system dynamics and the ability to predict future events.

In most applications, observations occur at random, unevenly spaced and unaligned time stamps. In such a setting we therefore consider two underlying processes
$
\left(X_{t}\right)_{t\in\mathbb{R}}
$
and
$
\left(Y_{t}\right)_{t\in\mathbb{R}}
$
that are only observed at discrete and finite timestamps in the form of two collections of data points:
$
\left(x_{t_x}\right)_{t_x \in I_x},\left(y_{t_y}\right)_{t_y \in I_y}
$. 
We adapt our definition of causality to this different theoretical framework. Let $\phi$ be a causal convolution kernel with delay $\tau$ (i.e. $\phi_{\tau}(t) = 0$ whenever $t < \tau$), $(X)$ and $(Y)$ two continuous time stochastic
processes, for instance, two Wiener processes.
A popular instance of a causation kernel is for instance the exponential kernel:
$
\phi(t)_{\tau, \Theta = \left(\alpha, \beta\right)} = \alpha \exp( - \beta (t - \tau) ) \text{ if } t > \tau, 0 \text{ otherwise}.
$
We assume for instance that $(X_t)$ is an Ornstein-Uhlenbeck process or a Brownian motion (Wiener process) \cite{karatzas2012brownian} and
\begin{equation}
dY_t = dW^Y_t + \int_{s \in \mathbb{R}} \phi(s)_{\tau, \Theta} dX_{(t-s)}
\label{eq:conv}
\end{equation}
where $W^Y$ and $W^X$ are two independent Brownian motions whose increments are classically referred to as the \emph{innovation process}. In the following, the parameter $\tau$ will be referred to as \emph{lag}. We will consider that $(Y)$ is lagging with characteristic delay $\tau$ behind $(X)$ which is causing it.

We adopt the cross-correlogram based causality estimation approach developed in \cite{huth2014high},
in order to be consistent with Granger's definition of causality as linear predictive ability of $(dX_{s<t})$ and $(dY_{s<t})$ for the random variable $dX_t$ \cite{granger1969investigating}. 

Let $(X)$ and $(Y)$ be two Wiener processes. 
We consider that $(X)$ has a causal effect on $(Y)$ if 
$(dX_{s <t})$ is a more accurate linear predictor of $dY_t$ 
in square norm error than $(dY_{s < t})$ is an accurate linear predictor of $dX_t$.
In other words $(X)$ \emph{causes} $(Y)$ if and only if
\begin{align}
    & E \left[ \left( dX_t - E(dX_t | dY_s, s < t) \right)^2 \right] > \nonumber \\
    & \hspace{1cm} E \left[ \left( dY_t - E(dY_t | dX_s, s < t) \right)^2 \right].
\end{align}
In order to quantify the magnitude of this statistical causation, Huth and Abergel introduced in \cite{huth2014high} the Lead-Lag Ratio (LLR) between $\left(X\right)$ and $\left(Y\right)$
as 
\begin{equation}
LLR_{X \Rightarrow Y}
=
\frac{\sum_{h>0}\rho_{XY}^{2}\left(h\right)}{\sum_{h<0}\rho_{XY}^{2}\left(h\right)} \label{eqn:llr}
\end{equation}
where $\rho_{XY}(\cdot)$ is the cross-correlation between the second order stationary processes $(X)$ and $(Y)$.
The analysis conducted in \cite{huth2014high} proved $(X)$ \emph{causes} $(Y)$ is equivalent to
\begin{equation*}
    LLR_{X \Rightarrow Y} < 1
\end{equation*}
thereby yielding an indicator of causation intensity between processes which depends $\phi_{\tau, \Theta}$ through (\ref{eq:conv}). 
%The very definition of the model paves the way to a frequentist linear second-order statistical analysis. The approach presented here is therefore different from Bayesian analysis by discrete state-space models \cite{durbin2012time} which is not directly applicable to this framework. Discrete state-space models would indeed only consider a theoretical process defined over $\mathbb{Z}$ as opposed to $\mathbb{R}$ for the underlying continuous processes we consider here.

\subsection{Challenges with real world data:}
Unfortunately, in practical applications, time series data sets often present three main challenges that hinder the estimation of even linear causal dependencies:
\begin{itemize}
\lessdenselist
\item \textbf{Irregular Sampling:} Observations are collected at irregular intervals both within and across processes complicating the application of standard causal inference techniques that rely on evenly spaced timestamps that align across processes.
\item \textbf{Long Range Dependencies (LRD):} Long range dependencies can result in increased and non vanishing variance in correlation estimates.
\item \textbf{Scale:} Real-world time series are often very large and high dimensional and are therefore often stored in distributed fashion and require communication over limited channels to process.
% and stored in an inherently decentralized manner on a distributed file system and requiring coordination, typically over limited communication channels. 
\end{itemize}

In the following we show as in \cite{huth2014high} that naive interpolation of irregularly sampled data may yields spurious causality inference measurements. 
We also prove that eliminating LRD is crucial in order to obtain consistent correlation estimates.
Unfortunately, standard time domain LRD erasure requires sorting the data chronologically and is therefore costly in the distributed setting. 
These costs are further exacerbated by time domain fractional differentiation which scales quadratically with the numbers of samples. 
% Unfortunately time domain erasure of LRD requires order-ing the data chronologically which may require communi-cating large quantities of records as fractional differentiationscales quadratically with the number of samples.

To address these three critical challenge we propose a Fourier transform based approach to causal inference.
%Spectral analysis of stochastic processes has
%focused on discrete time processes with aligned timestamps
%\cite{Brockwell:1986:TST:17326} leveraging the speed of the Fast
%Fourier Transform (FFT) algorithm. This procedure indeed requires samples
%sorted in chronological order and observed with regular intervals.
%Our key contribution is to roll back this procedure to its fundamental mathematical
%principle.
Projecting on a Fourier basis can be done with a simple sum operator
for irregularly sampled data as described in \cite{parzen2012time}.
%She show that the supplementary cost incurred by leaving the divide and conquer FFT approach 
%is largely compensated by the generality of the resulting method and its
%many computational advantages. 
A novel and salient byproduct of our estimation technique is that there is no need to
sort the data chronologically or gather the data of different sensors
on the same computing node. We use Fourier transforms
as a signal compressing representation where cross-correlations and
causal dependencies can be estimated with sound statistical methods
all while minimizing memory and communication overhead.
% in a memory and avoiding communication. 
In contrast to sub-sampling in which aliasing obscures short-range interactions, 
our methods does not introduce aliasing enabling the study of sort-range interactions.
An exciting aspect of compressing by Fourier transforms is that it only affects the variance of the cross-correlogram without destroying the opportunity to study inter-dependencies at a small time scale.
In section 2 we show that leveraging the frequency domain representation we present communication avoiding consistent spectral estimators \cite{brillinger1981time} for cross-dependencies.
We first compress the time series by projecting \emph{without interpolation or reordering} directly onto a reduced Fourier basis, thereby locally compressing the data. 
Spectral estimation then occurs in the \emph{frequency domain}
prior to being translated back into the time domain with an inverse Fourier transform. 
The resulting output can be used to compute unbiased Lead-Lag Ratios and thereby identify statistical causation. 

% in the presence of irregularly sampled data. % and LRD. % <- Is that (the LRD statement) correct? you cover LRD next
%No interpolation or ordering of the data is needed which is new to the best of our knowledge for causality estimation.

In section 3, we provide a method to approximately erase LRD in the frequency domain, which has tremendous computational advantages as opposed to time domain based methods. 
Our analysis of LRD erasure as fractional pole elimination in frequency domain guarantees the causal estimates we obtain are not spurious unlike those calculated naively on LRD processes \cite{doukhan2003theory,granger1988causality}.
Finally, we apply these methods to synthetic data and several terabytes of real financial market trade tables. 

In section 4, we present a novel analysis of the trade-off between estimator variance and communication bandwidth which precisely assesses the cost of compressing time series prior to analyzing them.
A three-fold analysis establishes the statistical soundness of the contributions that address the three issues mentioned above. Studying data on compressed representations comes at an expected cost. In our setting this supplementary variance can be decreased in an iterative manner and with bounded memory cost on a single machine. These properties cannot be replicated to the best of our knowledge by time domain based sub-sampling.

\vspace{-0.2cm}

\section{Interpolation and spurious lead-lag}

%Let us consider a collection of time series with irregularly sampled
%records 
%$
%\left(
%\left(x_{t_{1}}^{1}\right)_{t_{1}\in %I_{1}},\ldots,\left(x_{t_{d}}^{d}\right)_{t_{d}\in I_{d}}
%\right)
%$.
%The theoretical underlying stochastic processes in continuous time
%$
%\left(
%\left(X_{t}^{1},\ldots,X_{t}^{d}\right)_{t\in\mathbb{R}}^{T}
%\right)
%$
%is typically a multivariate fractional Brownian motion, exponential
%of a Brownian motion or Ornstein-Ulhenbeck process \cite{karatzas2012brownian}. 

In this section, we first review existing techniques for interpolated time-domain estimation of second-order statistics in the context of sparse and random sampling  along the time axis. Interpolating data is a usual
solution in order to be able to use classic time series analysis \cite{parzen2012time,wiener1949extrapolation,friedman1962interpolation,linsay1991efficient}. Unfortunately it is not always suitable,
as it can create spurious causality estimates and implies a supplementary
memory burden.

\vspace{-0.2cm}

\subsection{Second order statistics and interpolated data}
In order to infer a linear model from cross-correlogram estimates by solving the Yule-Walker equations \cite{Brockwell:1986:TST:17326} or to compute a LLR (\eqnref{eqn:llr})
one needs to estimate the cross-correlation structure of two time series.
Let $\left(X\right)$ and $\left(Y\right)$ be two centered stochastic
processes whose cross-covariance structure is stationary:
\begin{equation}
\gamma_{XY}\left(h\right)=E\left(X_{t-h}Y_{t}\right).
\end{equation}
If data is sampled regularly
$
\left(x_{n\Delta t}, y_{n\Delta t}\right)_{n = 0 \ldots N-1}
$
a consistent estimator for
$
\gamma_{XY}\left(h\right)
$
is:
\begin{equation}
\widehat{\gamma_{xy}\left(h\right)}=\frac{1}{N-h-1}\sum_{n=h}^{N-1}x_{\left(n-h\right)\Delta t}y_{n\Delta t}
\label{eqn:crosscov}
\end{equation}
(we use $\widehat{A}$ to denote an estimator for $A$).
Classically, cross-correlation estimates can subsequently be computed as
\begin{equation}
\widehat{\rho_{XY}\left(h\right)}=
\frac{
    \widehat{\gamma_{XY}\left(h\right)}
    }
{
\sqrt{\widehat{\gamma_{XX}\left(0\right)}\widehat{\gamma_{YY}\left(0\right)}}
}
\label{eqn:crosscor}
\end{equation}
using any consistent cross-covariance estimator. % $\widehat{\gamma}$.

\par{\textbf{Interpolating irregular records:}}

The standard consistent estimator \eqnref{eqn:crosscov} cannot be computed when $(x)$ and $(y)$ do not share common timestamps.
% for unaligned time stamps
% and therefore is no longer applicable whenever observed processes.
% The standard consistent estimator \eqnref{eqn:crosscov} cannot be computed for unaligned time stamps
% and therefore is no longer applicable whenever observed processes $(x)$ and $(y)$ do not share common timestamps.
A classical way to circumvent the irregular sampling issue is therefore
to interpolate the records 
$
\left(x_{t_{x}}\right)_{t_{x}\in I_{x}}
$
and
$
\left(y_{t_{y}}\right)_{t_{y}\in I_{y}}
$
onto the set of timestamps 
$
\left(n\Delta t\right)_{n = 0 \ldots N-1}
$
therefore yielding two approximations 
$
\left(\widetilde{x_{n \Delta t}}\right)_{n = 0 \ldots N-1 }
$
and
$
\left(\widetilde{y_{n \Delta t}}\right)_{n = 0\ldots N-1 }
$
that can be studied as a synchronous multivariate time series. 
An adapted cross-covariance estimate is then 
$
\widehat{\widetilde{\gamma_{xy}\left(h\right)}}
=
\frac{1}{N-h-1}\sum_{n=h}^{N-1}\widetilde{x_{\left(n-h\right)\Delta t}}\widetilde{y_{n\Delta t}}.
$
While there are many interpolation techniques, a commonly used method is \emph{last observation carried forward} (LOCF).
Note that interpolation may require substantial additional memory to render each time series at the resolution of interactions which can be millisecond scale in many crucial applications such as studying stock market interactions.

% Interpolation can be accomplished using a range of techniques (e.g.,  ``last observation carried forward'' (LOCF).
We now consider the causality inference framework introduced in
\cite{huth2014high} and show how the LOCF
interpolation technique creates spurious causality estimates.

\par{\textbf{Bias in LLR with irregularly sampled data:}}
The $LLR$ can be computed by several methods. Cross-correlation
measurements on a symmetric centered interval are sufficient statistics for this estimator. Therefore one can use synchronous cross-correlation estimates
on interpolated data in order to compute the $LLR$. Carrying the last observation forward (LOCF) has been proven to create a bias in lag estimation in \cite{huth2014high}. 
% This interpolation method is spurious % <- what does this mean ok
% as it 
The LOCF interpolation method introduces a causality estimation bias in which a process sampled at a higher frequency will be seen as causing another process which is sampled less frequently although these correspond to Brownian motions with simultaneously correlated increments.

\vspace{-0.2cm}

\subsection{Interpolation-free causality assessment}
The \emph{Hayashi-Yoshida} (HY) estimator was introduced in \cite{hayashi2005covariance} to address this spurious causality estimation issue. 
The HY estimator of cross-correlation does not require data interpolation and has
been proven to be consistent with processes sampled on a quantized
grid of values \cite{hoffmann2013estimation} in the context of High Frequency statistics in finance. 

\par{\textbf{Correlation of Brownian motions:}}
HY is adapted to measuring cross-correlations between irregularly
sampled Brownian motions. 
Considering the successor operator $s$ for the series of timestamps of a given process, let 
$\left[t, s(t)\right]_{t \in I_x}$
and 
$\left[t, s(t)\right]_{t \in I_y}$ 
be the set of intervals delimited by consecutive observations of $x$ and $y$ respectively.
The Hayashi-Yoshida covariance estimator over the covariation of $(X)$ and $(Y)$ \cite{karatzas2012brownian} is defined as 
\begin{equation}
\text{HY}_{[0, t]}(x, y) = \sum_{t \in I_x, t' \in I_y : \text{ov}(t, t')} (x_{s(t)} - x_t) \cdot (y_{s(t')} - y_{t'})    
\end{equation}
where $\text{ov}(t, t')$ is true if and only if $[t, s(t)]$ and $[t', s(t')]$ overlap. The estimator can be trivially normalized so as to yield a correlation estimate.

\par
\begin{figure}
    \centering
    \begin{tabular}{cc}
        \includegraphics[width=4cm]{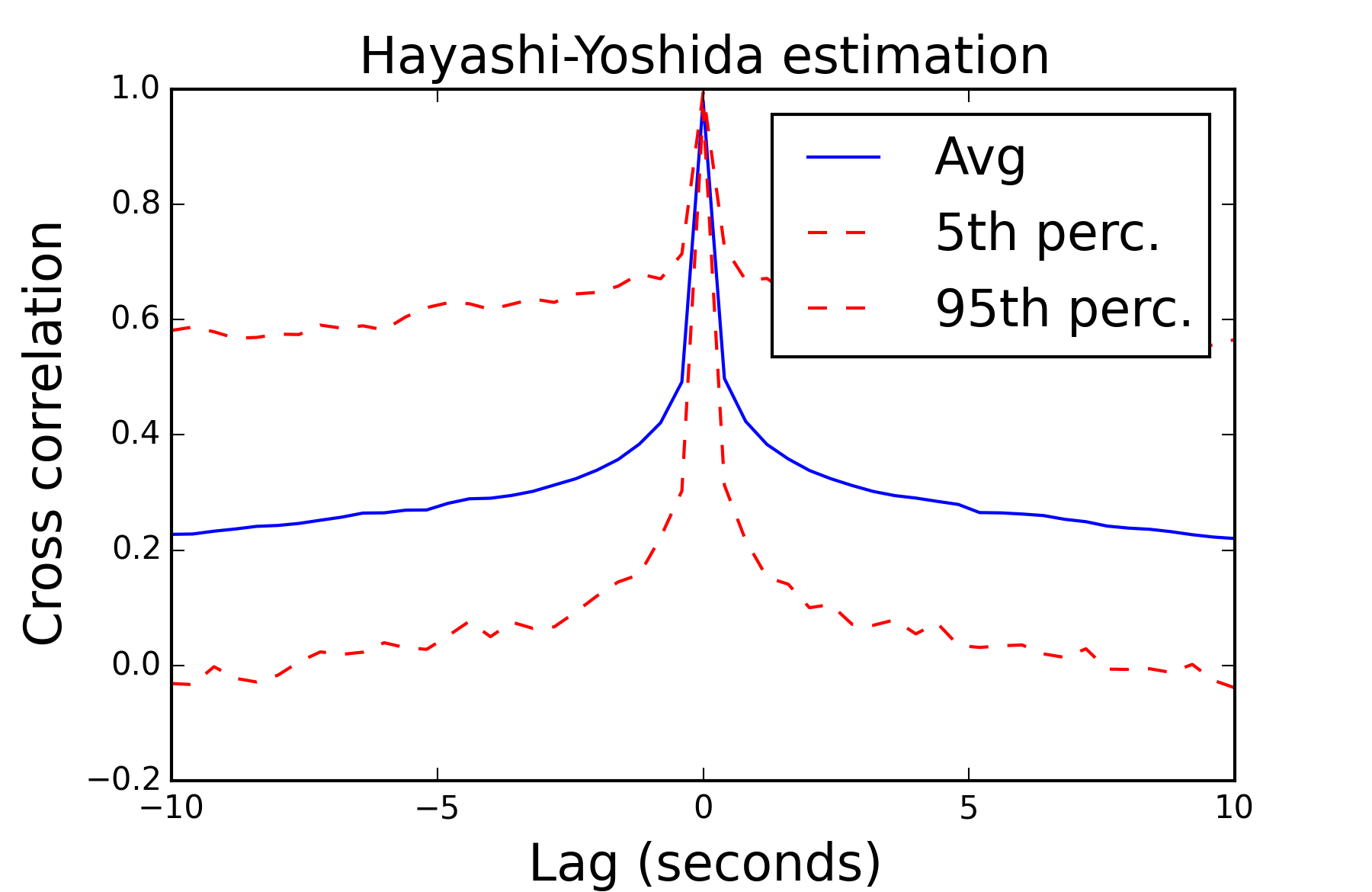} & \includegraphics[width=4cm]{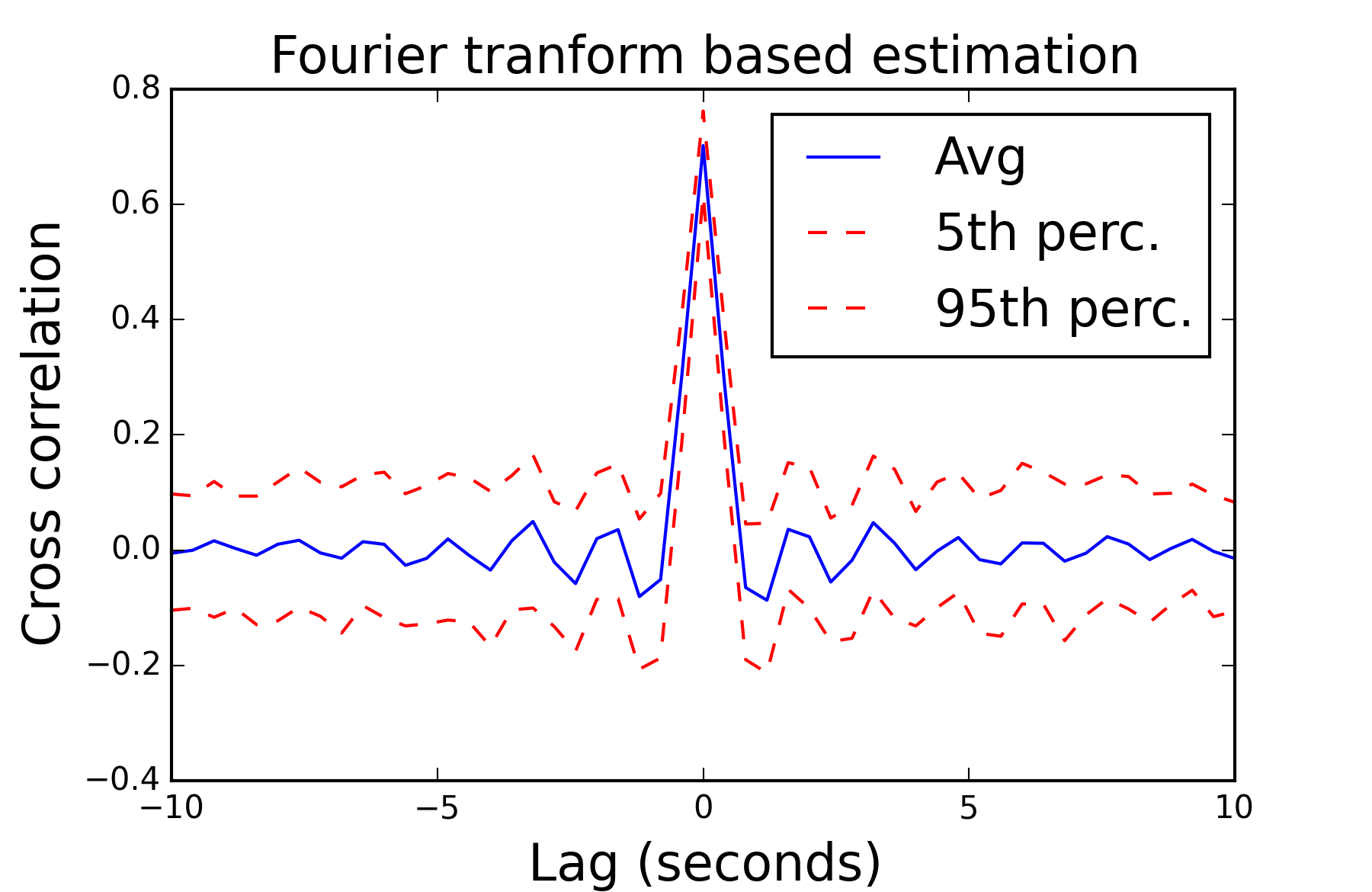}
    \end{tabular}
    \caption{
    \footnotesize{\textbf{LRD Erasure:}
    Monte Carlo simulation ($100$ samples) of two fractional Brownian motions with Hurst exponent $0.8$ and simultaneously correlated increments. Spurious slowly vanishing cross-correlation hinders the HY estimation but does not affect our estimation with LRD erasure (see Section 3) as evident by nearly zero cross-correlation for non-zero lag.
    }
    }
    \label{fig:HY_frac_diff}
    \vspace{-0.3cm}
\end{figure}

{\textbf{HY and fractional Brownian motions:}}
No interpolation is required with HY but unfortunately this estimator is only designed to handle full differentiation of standard Brownian motions. Figure \ref{fig:HY_frac_diff} shows how HY fails to estimate cross-correlation of increments on a fractional Brownian motion whereas the technique we present succeeds.
In the following, we show how our frequency domain based analysis naturally handles irregular observations and is able to fractionally differentiate the underlying continuous time process. This is in particular
necessary when one studies factional Brownian motions with correlated increments. In the interest of concision, we refer the reader to \cite{flandrin1989spectrum} for the definition of a fractional Brownian motion.

\subsection{Fourier transforms for irregularly sampled data}
Our alternative approach to estimating cross-correlograms is based on the definition of the Fourier transform of a stochastic process.
Considering a continuous time stochastic process $\left(X_t\right)_{t \in [0 \ldots T]}$ and a frequency
$f\in\left[0\ldots2\pi\right]$, the Fourier projection of $\left(X\right)$
for the frequency $f$ is defined as 
\begin{equation}
P_{f}\left(X\right)=\int_{t = 0}^{T}X_{t}e^{-ift}dt
\end{equation}
where $i$ is the imaginary number.
Much attention has been focused on the benefits of the FFT algorithm which has been designed for the very particular base of ordered and regularly sampled observations. 
\emph{Our key insight is to go back to the very definition of the Fourier transform as an integral and express it empirically in summation form~\cite{brillinger1981time,parzen2012time}.}
Moreover, if the process $\left(X\right)$ is observed
at times $\left(t_{1},\ldots,t_{N}\right)$, one can estimate the Fourier
projection by 
\begin{equation}
\widehat{P_{f}}\left(x\right)=\sum_{n=1}^{N}x_{t_{n}}e^{-ift_{n}}.
\end{equation}
Therefore we propose the following simple framework for frequency domain based linear causal inference:
\begin{enumerate}
    \lessdenselist
    \item Project $(x)$ and $(y)$ on to a \emph{reduced} Fourier basis.
    \item Estimate the cross-spectrum of $(X)$ and $(Y)$ in the frequency domain.
    \item Apply the inverse Fourier transform to the cross-spectrum to recover the cross-correlogram and infer the linear causal structure.
\end{enumerate}

The intuition behind this estimation method is a change of basis
that allows us to compute cross-covariance estimates without needing to address the irregularity of timestamps. 
% any concerns
% about the characteristics of the timestamps.
Indeed the
power spectrum $f(\cdot)$ is the element-wise Fourier transform of 
$$
\gamma\left(\cdot\right)=\left[\begin{array}{cc}
\gamma_{XX}\left(\cdot\right) & \gamma_{YX}\left(\cdot\right)\\
\gamma_{XY}\left(\cdot\right) & \gamma_{YY}\left(\cdot\right)
\end{array}\right].
$$
Therefore, in order to estimate this function one may infer what corresponds to its frequency domain representation and then compute the inverse Fourier transform of the result.

\textbf{Projecting onto Reduced Fourier Basis:} 
We first project $\left(X\right)$ and $\left(Y\right)$
onto the elements of the Fourier basis of frequencies
$
\left(l \Delta f \right)_{l = 0 \ldots P}
$,
namely the pair
$
\left(
    P_{l \Delta f}\left(X\right)
\right)_{l = 0 \ldots P}
$
and 
$
\left(
    P_{l \Delta f}\left(Y\right)
\right)_{l = 0 \ldots P}
$. 
By projecting onto a single relatively small set of orthonormal functions, we are able to compress and effectively re-align the observations $\left(x\right)$ and $\left(y\right)$. 
In practice using only a few thousand basis functions we are able to accurately recover the cross-correlogram.
Finally, this computation is sufficiently fast to execute interactively on a single laptop and can be easily expressed using the map-reduce framework.

% only requires a few lines of code and can be expressed trivially in the map-reduce framework.

\textbf{Estimating the Cross-spectra:}
Computing projections onto a reduce Fourier basis enables exploratory data analysis through the
study of the cross-spectrum of $\left(X\right)$ and $\left(Y\right)$
\begin{equation}
\left(I_{XY}\left(l \Delta f\right)\right)_{l = 0 \ldots P} = \left(P_{l \Delta f}\left(X\right)\times\overline{P_{l \Delta f}\left(Y\right)}\right)_{l = 0 \ldots P}. \label{eqn:crossspectrum}
\end{equation}
An \emph{inconsistent} estimator for the cross-spectrum is:
\begin{equation}
\left(\widehat{I_{XY}\left(l \Delta f\right)}\right)_{l = 0 \ldots P}
=
\left(
\widehat{P_{l \Delta f}\left(x\right)}
\times
\overline{
\widehat{P_{l \Delta f}\left(y\right)}
}
\right)_{l = 0\ldots P}    .
\label{eqn:crossspectrumest}
\end{equation}
%In frequency domain, 
Local averaging of \eqnref{eqn:crossspectrumest} with respect to frequencies is widely used \cite{brillinger1981time,Brockwell:1986:TST:17326,parzen2012time} in cross-spectral analysis to identify the characteristic frequencies at which stochastic processes interact although they are observed at irregular times.
%Computing the inverse Fourier transform of \eqnref{eqn:crossspectrumest} yields a consistent estimator for cross-correlation.
Unfortunately, to compute characteristics delays or LLR (crucial steps in linear causal inference) we still need to estimate the cross-correlogram.

% We obtain a consistent estimator of the cross-correlogram by applying the inverse Fourier transform to the cross-spectrum \eqnref{eqn:crossspectrum}. %is that better?
% \emph{As a consequence we are able to study the characteristic delays and LLR and therefore infer linear causal structure without the need for interpolation or data reordering.}

\par{\textbf{Estimating the Cross-correlogram:}}
To estimate the cross-correlogram we can take the inverse Fourier
transform of the cross-spectrum 
$
\left(I_{XY}\left(l \Delta f\right)\right)_{l = 0 \ldots P}
$
%given by the following equation:
which translates frequency analysis back into the time domain:
%The following equation relates cross-spectrum and cross-covariance:
\begin{equation}
\gamma_{XY}^{P}\left(h\right)=\frac{1}{P}\sum_{l=0}^{P}I_{XY}\left(l \Delta f\right)e^{i l \Delta f h}.
\end{equation}

Using the following \emph{consistent} estimator: 
\begin{equation}
\widehat{\gamma_{XY}^{P}}\left(h\right)=\frac{1}{P}\sum_{l=0}^{P}\widehat{I_{xy}}\left(l \Delta f\right)e^{i l \Delta f h}
\end{equation}
of the cross-covariance we can directly compute a \emph{consistent} estimator of the cross-correlation using equation \eqnref{eqn:crosscor}.
% Consider two processes $\left(X\right)$ and $\left(Y\right)$
% observed with two different finite sets of irregularly sampled timestamps
% $I_{x}\subset\mathbb{R}$ and $I_{y}\subset\mathbb{R}$. 
The cross-correlation
between $(X)$ and $(Y)$ can now be estimated in the time domain with a discrete grid $G_h$ of lag values 
ranging from $-L \Delta h$ to $L \Delta h$ with a resolution $\Delta h$.
As expected, aliasing will occur if the user specifies a resolution in the cross-correlation estimate that is much higher than the average sampling frequency of the time series \cite{parzen2012time}.

% is then obtained as
% and then a consistent estimator of the cross-correlation using equation \eqnref{eqn:crosscor}
% which normalizes measurements.

%\joey{You could drop this paragraph is you really needed space}
%\par{\textbf{Equivalence between time and frequency domain:}}
In contrast to more cumbersome time domain synchronization relying on interpolation based methods (LOCF) or interval matching based estimations (HY), our method elegantly addresses time synchronization in the \emph{frequency domain}.
While earlier work \cite{parzen2012time,brillinger1981time} has considered the application of frequency domain analytics to irregularly sampled data, our method is the first to translate back to the time domain to recover a consistent estimator of correlation. 
Alternatively, Lomb-Scargle periodogram \cite{scargle1982studies,lomb1976least} also enables the frequency domain analysis of irregularly observed data but suffers from the supplementary cost of a least square regression.
To the best of our knowledge we are the first to use frequency domain projections to compute the cross-correlogram in order to infer linear causal structure.

\vspace{-0.4cm}

\subsection{The statistical cost of compression}
%This previous observation enables the analysis in time domain via estimation in the frequency domain.
Central to the communication and memory performance of our technique is the ability to use a small number of Fourier projections relative to the number of observations and still accurately recover the cross-correlogram.

\par{\textbf{Cross-correlogram Estimator Consistency:}}
We can characterize the statistical properties of the cross-spectral estimator \cite{brillinger1981time,parzen2012time,Brockwell:1986:TST:17326}.
In particular, it is well known that for two distinct non-zero frequencies $f_1$ and $f_2$ the estimators $\widehat{I_{XY}(f_1)}$ and $\widehat{I_{XY}(f_2)}$ are asymptotically independent. 
Consequently, to obtain an estimator with variance $O(V)$ the user will need to project on $\frac{1}{V}$ frequencies. 
We confirm this result numerically in Figure \ref{fig:emp_causation}.
The element-wise product of Fourier transforms is converted into the time domain by the inverse Fourier transform to yield a cross-correlogram. 
With very large datasets in which $N >> \frac{1}{V}$ we obtain the suitable compression property of our algorithm.

\par{\textbf{Issues with Non-smooth Cross-correlograms:}}
As expected, deterministic lags or seasonal components
can result in
%issues as the cross-correlation function is non-smooth in that case. 
Fourier compression artifacts in the inverse Fourier transform. 
However, statistical estimation and removal of these deterministic components is standard in time series analysis \cite{Brockwell:1986:TST:17326,brillinger1981time}.
In the context of estimating non-trivial stochastic causal relationships 
(e.g., social networks, pairs of stock prices the financial markets, cyberphysical systems)
random perturbations affect the causation delay. 
In these settings, the theoretical cross-correlation function is smooth.
As a consequence, a few Fourier projections suffice to accurately represent the cross-correlogram in frequency domain.

\subsection{Example of time domain exploratory data analysis through the frequency domain}
%The novelty of our approach therefore resides in expressing time domain
%analysis in terms of its frequency domain counterpart which is advantageous
%in terms of signal processing and computation burden.
The time domain exploratory analysis we enable makes lead-lag relationships self-explanatory as shown
in Figure~\ref{fig:Illustration-cross-corre}. We show in the following that it is not hindered
by biases related to the fact that one process is sampled more seldom than the other.

\par
\begin{figure}
\begin{centering}
\begin{tabular}{cc}
LOCF cross-correlation & Fourier cross-correlation \\
\includegraphics[width=3.8cm, trim={0.60cm 0cm 1.0cm 0.25cm}]{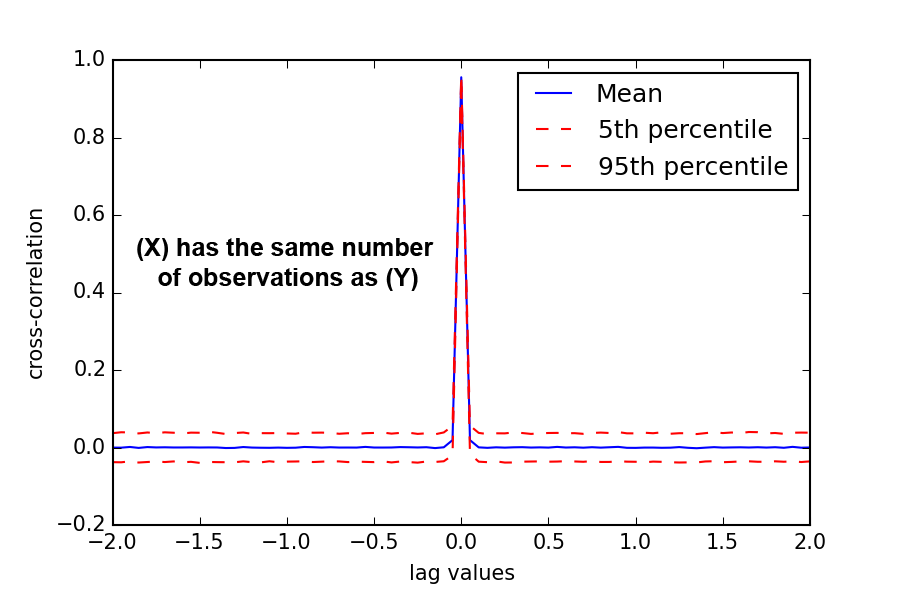} & \includegraphics[width=3.8cm, trim={0.60cm 0cm 1.0cm 0.25cm}]{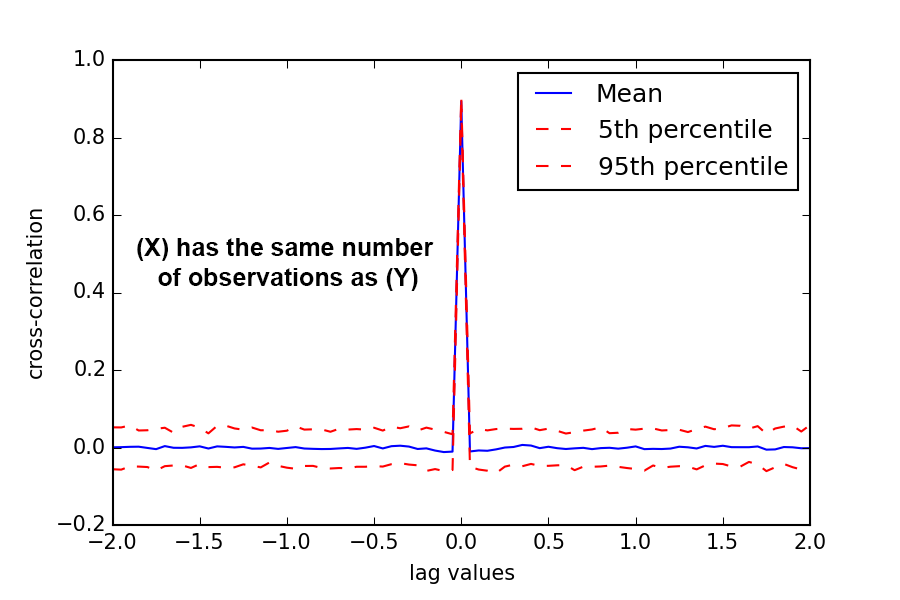} \\
\includegraphics[width=3.8cm, trim={0.60cm 0cm 1.0cm 0.25cm}]{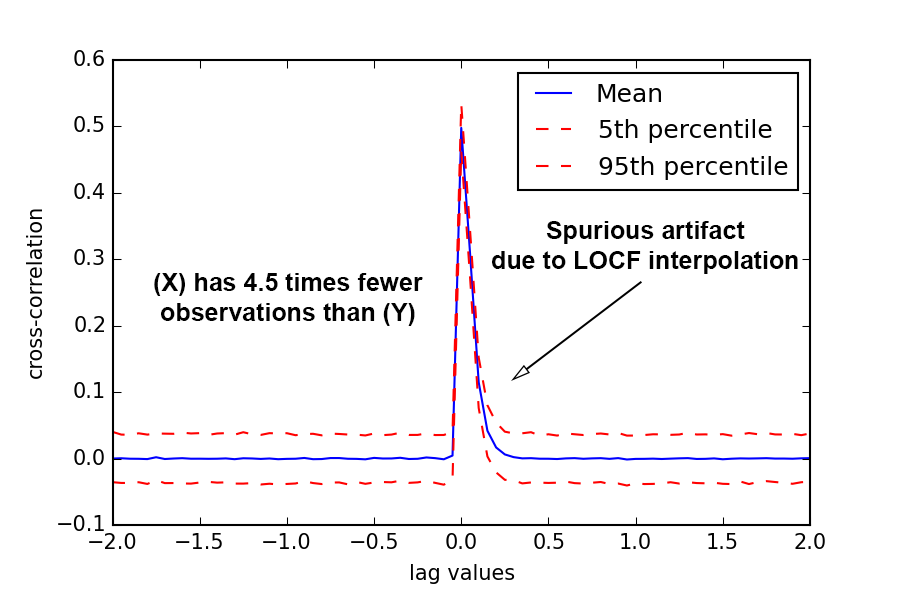} & \includegraphics[width=3.8cm, trim={0.60cm 0cm 1.0cm 0.25cm}]{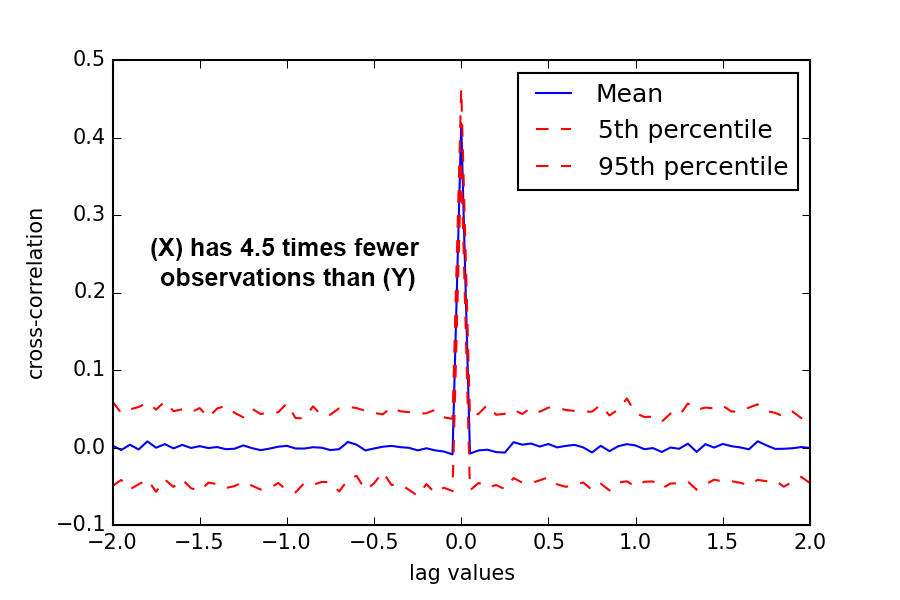} \\
\includegraphics[width=3.8cm, trim={0.60cm 0cm 1.0cm 0.25cm}]{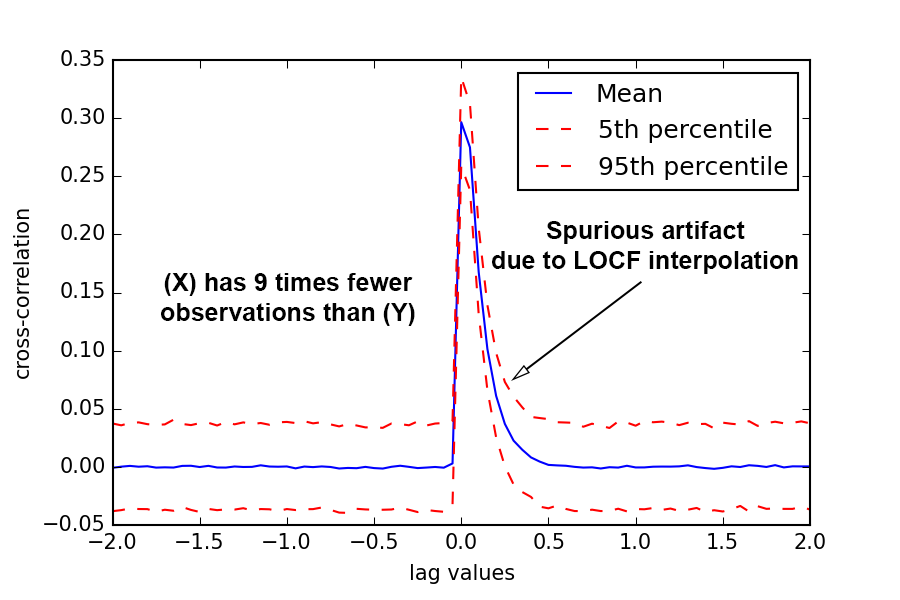} & \includegraphics[width=3.8cm, trim={0.60cm 0cm 1.0cm 0.25cm}]{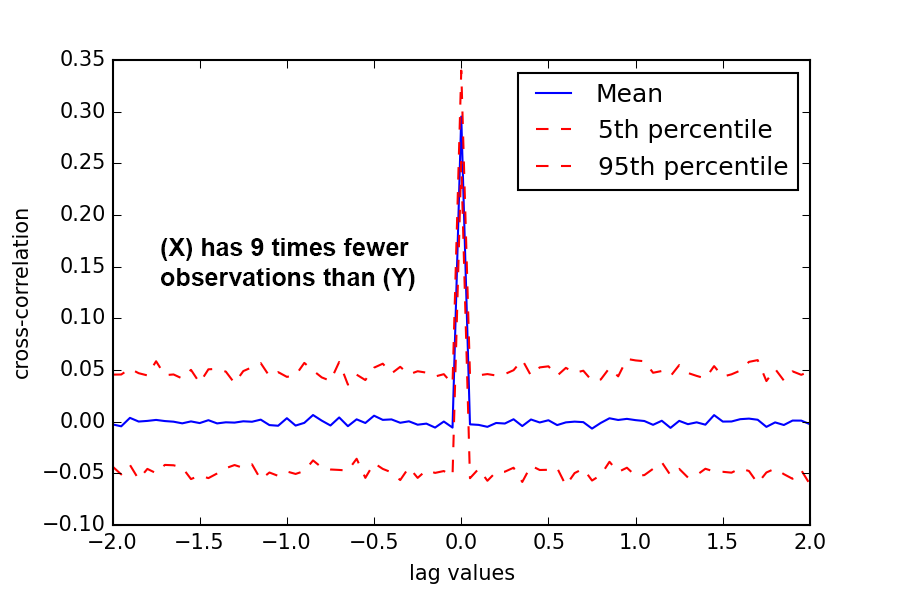} \\
\end{tabular}
\par\end{centering}
\caption{\footnotesize{Cross-correlograms of
LOCF interpolated data versus estimation via compression in frequency domain.
The latter estimate does not present any spurious asymmetry due to the uneven
sampling frequencies}}
\label{fig:spurious_caus_2}
\end{figure}

{\textbf{Numerical assessment of frequency domain based correlation measurements:}}
We demonstrate, through simulation, that the spurious causation issue that plagues the LOCF interpolation~\cite{huth2014high} does not appear in our proposed method. 
We consider two synthetic correlated Brownian motions that do not feature any lead-lag and compare the estimation of LLR provided by two time domain interpolation methods and our approach. 
After having sampled these at random timestamps, in Table \ref{fig:spurious_caus_1} and Figure \ref{fig:spurious_caus_2} we compare
the cross-correlation and LLR estimates obtained by LOCF interpolation and our proposed frequency domain analysis technique confirming that our method does not introduce spurious causal estimation bias.
% eh good enough.... 
% Alright!
% This confirms our work-flow does not have any spurious LLR issue as opposed to the LOCF method.

\begin{table}[h]
\begin{centering}
\begin{tabular}{c|c|c}
$\frac{N_{1}}{N_{2}}$ & LOCF interpolation LLR & Fourier transform LLR\\
& Avg +- std & Avg +- std\\
\hline 
$1$ & 
$0.998 +- 0.135$ & $1.021 +- 0.166$\\
\hline 
$4.5$ & $6.863 +- 1.678$ & $1.053 +- 0.320$\\
\hline 
$10$ & $7.277 +- 1.854$, & $1.107 +- 0.391$\\
\hline 
\end{tabular}
\par\end{centering}
\caption{
\footnotesize{
Comparison of LLR ratios with LOCF and Fourier transforms ($1000$ projections) for simultaneously correlated Brownian motions with different sampling frequencies. The LLR ratios should all be $1$, one can observe the bias in the LOCF method.
}
}
\label{fig:spurious_caus_1}
\end{table}

\vspace{-0.4cm}

\section{The Long Range Dependence (LRD) Issue}

A stochastic process is said to be long range dependent if it features cross-correlation magnitudes whose sum is infinite \cite{doukhan2003theory}. Many issues arise in that case with
correlation estimates becoming spurious. This phenomenon was
first discover when Granger studied the concept of cointegration between
Brownian motions (integrated time series) \cite{granger1988causality}. On sorted Brownian motion
data, this effect can be addressed by differentiating the time series,
namely computing 
$
\left(\Delta X_{t}\right)_{t\in\mathbb{Z}}=\left(X_{t}-X_{t-1}\right)_{t\in\mathbb{Z}}
$.
For fractional Brownian motion and LRD time series,
the fractional differentiation operator needs to be computed. It is
defined as 
\begin{equation}
    \left(\Delta^{\alpha}X_{t}\right)_{t\in\mathbb{Z}}
    =\left(\sum_{h=0}^{\infty}\frac{\prod_{j=0}^{h-1}\left(\alpha-j\right)\left(-X_{t-h}\right)^{h}}{h!}\right)_{t\in\mathbb{Z}}.
    \label{eq:fracdiff}
\end{equation}
Therefore, to study the cross-correlation structure of two integrated
or fractionally integrated time series, one would have to compute
$
\left(\Delta X_{t}\right)_{t\in\mathbb{Z}}
$ or 
$
\left(\Delta^{\alpha}X_{t}\right)_{t\in\mathbb{Z}}
$.
The latter requires chronologically sorted data and synchronous timestamps and has
a quadratic time complexity with respect to the number of samples.

\subsection{Erasing memory in the frequency domain}

Erasing memory is of prime importance, in the case of the study of
Brownian motions and fractional Brownian motions alike. As
pointed out in \cite{doukhan2003theory}, LRD arises in many systems,
in particular those managed by humans, because of their ability to
learn from previous events and therefore keep of memory of these in
their future actions. From a computational and statistical point of
view, it is challenging to erase.

\par{\textbf{Equivalence between differentiation in time domain and element-wise multiplication in frequency domain:}}
Let $\left(X_{t}\right)_{t\in\left[0,T\right]}$ be a continuous process
whose fractional differentiate of degree $\alpha$, $d^{\alpha}X$ is Lebesgue-integrable with probability $1$.
If $X_{t}$ vanishes at the boundaries of the interval, classically,
almost surely,
$$
P_{f}\left(d^{\alpha}X\right)=\int_{t=0}^{T}e^{-ift}d^{\alpha}X_{t}=\left(if\right)^{\alpha}\int_{t=0}^{T}e^{-ift}X_{t}dt
$$
by a stochastic integration by part. 
Therefore, an estimate for 
$
P_{f}\left(d^{\alpha}x\right)
$
is 
$$
\widehat{P_{f}\left(d^{\alpha}X\right)}=\left(if\right)^{\alpha}\widehat{P_{f}\left(X\right)}.
$$

\par{\textbf{Erasing memory through fractional pole elimination:}}
The power spectrum of a fractional Brownian motion \cite{flandrin1989spectrum} with Hurst exponent
$H$ is asymptotically $~\frac{1}{f^{2H+1}}$ for $f << 1$. This is
the characteristic spectral signature of a long range dependent time
series. $H$ can therefore be estimated by the classical periodogram
method for an individual time series by conducting a linear regression
on the magnitude of the power spectrum about $0$ in a log/log scale
\cite{palma2007long}. Wavelets are another family of orthogonal basis
enabling a similar estimation \cite{percival2006wavelet}.
One can therefore see the fractional differentiation
operator of order $H+\nicefrac{1}{2}$ as a means to compensate for
a pole of order $2H+1$ in square magnitude in $0$. Multiplying the
Fourier transform of the signal by $\left(if\right)^{H+\nicefrac{1}{2}}$  eliminates the issue.
It does not require any preprocessing of the data,
no interpolation or re-ordering and we will show below that it has tremendous computational
advantages in the context of distributed computing in terms of communication avoidance.

%\subsubsection{Frequency domain zooming}
%Another way of thinking about the erasure of poles in that we apply %a high pass filter to both processes prior to computing the inner %product of their projections in Fourier domain. In order to %eliminate the detrimental impact of LRD on cross-correlation %estimation one magnifies high frequencies.
%A similar effect can be achieved by choosing the frequencies of the %projections as those that a priori are the most interesting to the %user. If the user is interested in interactions between processes %that characteristically occur with a short time span $\Delta t$ %then studying the spectral signature about $\frac{1}{\Delta t}$ and %then computing an inverse Fourier transform will select these very %interactions.

\par{\textbf{An approximation of differentiation in the case of discrete observations:}}
It is noteworthy though that this method is intrinsically approximate in the practical context of discrete
sampling. Indeed the multiplication rule for differentiation in frequency domain 
we proved in the context of stochastic processes does not directly apply in the context
of discrete observations. In order to ensure the soundness of the novel technique
we designed, we conduct several numerical experiments.

\vspace{-0.1cm}

\subsection{Testing frequency domain LRD erasure}

\begin{figure}
\centering
\includegraphics[width=7.5cm]{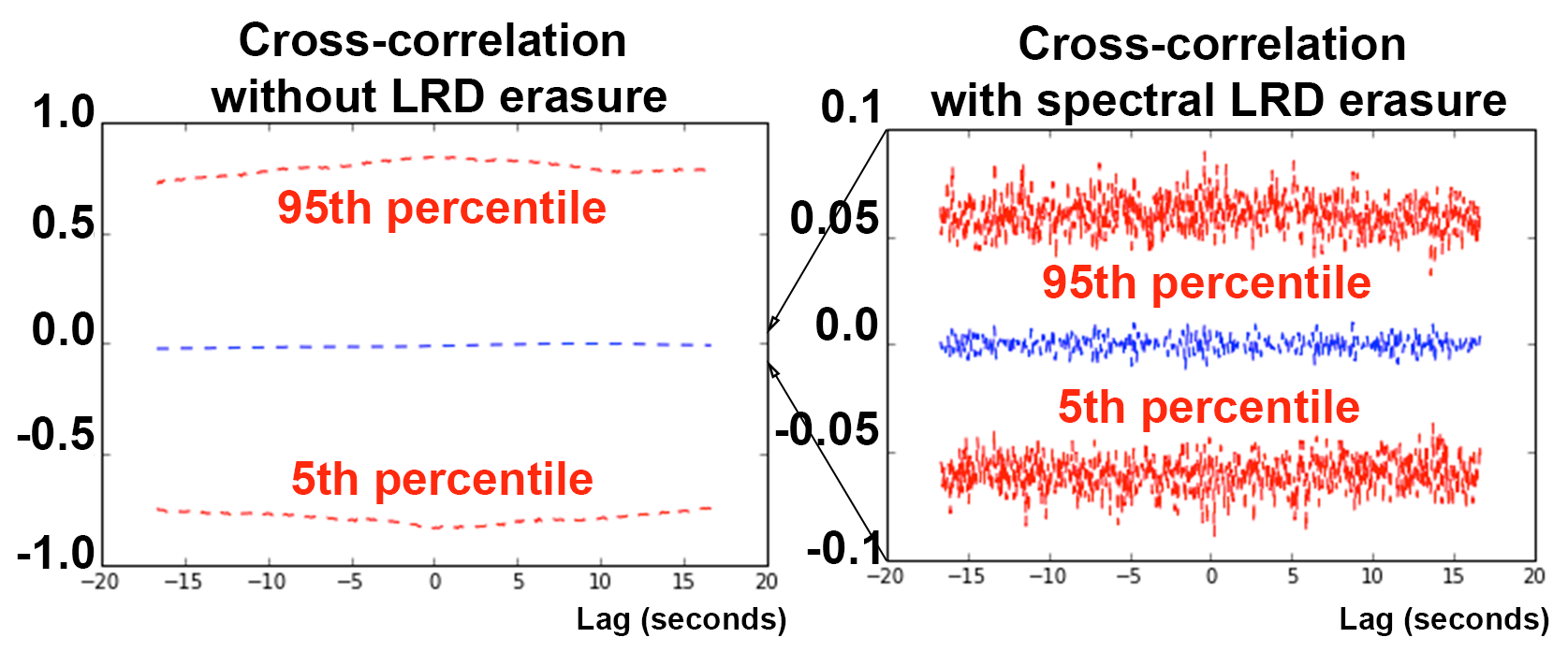}
\caption{
\footnotesize{
Spurious cross-correlation is erased by pole elimination in frequency
domain. The empirical cross-correlation distribution on the left is affected by high
magnitude spurious estimates. On the right, frequency domain fractional pole erasure eliminated the issue,
considerably narrowing down the interval between the $5^{\text{th}}$ and $95^{\text{th}}$ percentiles.
}
}
\vspace{-0.5cm}
\label{fig:LRDerasure}
\end{figure}

The example below considers two fractional Brownian motions $(X)$ and $(Y)$ Brownian motions with Hurst exponent $H=0.4$ \cite{doukhan2003theory}.
We compare the empirical distributions of cross-correlation estimates obtained over $100$ trials with and without
LRD erasure in frequency domain. In Figure \ref{fig:LRDerasure} we showcase an experiment with $9998$ uniformly random observations for $(X)$ and $6000$ uniformly random observations for $(Y)$.
While naive cross-correlation estimations lead to many spurious cross-correlation estimates with significantly high magnitudes
of estimated correlation values for processes that are in fact independent, 
($90\%$ of the empirical distribution between $-0.9$ and $0.9$)
the confidence interval we obtain with our
novel frequency domain erasure method by fractional pole elimination is narrower 
($90\%$ of the empirical distribution between $-0.05$ and $0.05$)
and enables reliable analysis.
The next section will expose the computational advantages 
of such a frequency domain based estimation as a communication avoidance
mechanism.

\vspace{-0.4cm}

\section{Distribution}

\vspace{-0.4cm}

Scalable computation is essential to practical causal inference in real-world \emph{big data} sets. 
Our proposed frequency domain approach provides a parallel communication avoiding mechanism to efficiently compress large time-series data sets while still enabling the estimation of cross-correlograms.

% address the challenges of scalable time series analytics by compressing the data in a highly parallel and unordered 
% Often data is distributed across several nodes in which case communication cost is often the dominant bottleneck. 
% Frequency domain compression is a communication avoiding algorithm here. 
% In this section we highlight the improvements it offers.

\vspace{-0.2cm}

\subsection{Computational setting}

Cluster computing presents the opportunity to enable faster analysis by leveraging the scale-out compute resources in modern data-centers. 
However to leverage scale-out cluster computing it is essential to minimize communication across the network as network latency and bandwidth can be orders of magnitude slower than RAM~\cite{peleg2000distributed,kumar1994introduction}.

\vspace{-0.2cm}

\subsection{Computational advantages}

One mechanism to distribute time domain analysis of time series is to construct overlapping blocks as described in \cite{belletti2015embarrassingly}. 
However, this technique only works if there is no LRD.
The need to specify the appropriate replication padding duration at preprocessing time makes it difficult to switch between the time scales at which cross-dependencies are computed.

The novel frequency domain based
methods we propose can entirely be expressed as trivial map-reduce aggregation operations
and do not require sorting or interpolating the data. 
Indeed, the use of projections on a subset of a Fourier basis only requires element-wise multiplication and then an aggregated sum to construct a unique concise signature in frequency domain for each time series that was observed. 
The amount of compression can be chosen by the user. This yields a flexible frequency domain probing method. Projecting on a few elements of the Fourier basis substantially reduces communication and memory complexity associated with the estimation of cross-correlograms.
As a consequence, the user can dynamically adjust the number of projections in order to progressively reduce the variance of the estimator.
%One may therefore start with coarse approximation and 

\subsection{Fourier compression as a communication avoidance algorithm}
The computation of Fourier projections is communication efficient in the distributed setting.
The Fourier projection can be calculated by locally computing the sum of the mapping of multiplications by complex exponentials. 
Then, only the local partial sums need to be transmitted across the network to compute the projections of the entire data set.
%therefore reducing the necessary bandwidth to a kilobyte size packet.
In this section, we study $d$ distinct processes with $N$ data points each. 
% By avoiding the Fast Fourier Transform algorithm, which requires sorted and aligned timestamps, our approach can be applied generically to study irregularly observed processes.
Let $V$ denote the desired variance for the cross-correlation estimator via the frequency domain.

%\subsubsection{Communication cost of aggregation with direct time domain covariance estimates}
%One can easily assess the needs in terms of packet sizes in the case of computing a distributed cross-correlation with an aggregation paradigm. After interpolation consider the sums
%$
%\left(\sum_{n=h}^{N-1}x_{\left(n-h\Deltat\right)\Delta t}y_{n}^{T}\right)_{h = 0 ... p}
%$
%and its computation on several machines each of which holds a subset of the data.
%In order to send as little data as possible one first aggregates the result of 
%a partial aggregation on each machine prior to sending these out on the network for
%the final aggregated output to be computed and communicated to the user.
%With $d$ data streams and a maximum lag $p$ chosen by the user, each machine will
%send out data of size $O(d^2 p)$. The analysis is equivalent if one chooses the HY %cross-correlation estimator.

\par{\textbf{Communication cost of aggregation with indirect frequency domain covariance estimates:}}
Now consider the set of Fourier projections
$
\left(
\widehat{P_{f}}\left(x\right)=
\sum_{n=1}^{N} x_{t_{n}} e^{-i f t_{n}}
\right)
_{f = 0, \Delta f, \ldots, P \Delta f}
$
which we aggregate on each single machine separately prior to sending them over the network.
The number of projections needed to have an estimator for cross-correlation with variance ~$V$ is $O(\frac{1}{V})$.
Therefore, the size of the message sent out by each machine over the communication medium is now $O(d \frac{1}{V})$ and representative of $O(d N)$ data points. 
If the user chooses $\frac{1}{V} \ll N$, our method effectively compresses the data prior to transmitting it over the network. 
%The resulting work-flow is illustrated in Figure \ref{fig:compression_diagram}.
It is noteworthy that the gain offered by this algorithm is system independent as long as communication between computing cores is the main bottleneck.

%\begin{figure}
%    \center
%    \includegraphics[height=10cm, trim = {1.5cm 0cm 0cm 0cm}, clip]{Fourier_compression}
%    \caption{Computing Fourier projections is a compression mechanism alleviating communication. Cross-spectra can be computed as well as fractional differentiation on compressed data stored on the driver. The result can then be translated in the time domain with an inverse Fast Fourier Transform which yields cross-correlation estimates highlighting causal dependencies.}
%    \label{fig:compression_diagram}
%\end{figure}

\par{\textbf{Distributed LRD erasure:}}
The computational complexity of fractional differentiation (\eqnref{eq:fracdiff}) is $O(N^2 d)$
in the time domain.
Furthermore, due to LRD, time domain fractional differentiation cannot be accomplished using the overlapping partitioning strategy proposed in \cite{belletti2015embarrassingly}.
% The operation of fractional differentiation as in equation \eqnref{eq:fracdiff} is computationally expensive.
% In time domain its computation is $O(N^2 d)$ in complexity 
% and cannot be made embarrassingly parallel because of the long range dependent structure which
% precludes the partitioning strategy proposed in \cite{belletti2015embarrassingly}.
Moreover, in distributed system, computing the fractional
differentiation of a signal would require transmitting the entire data set across the network.
As a consequence the bandwidth needed is $O(N d)$.
% In order to speed up computations on a cluster, one needs a different strategy.

Alternatively, fractional differentiation in the frequency domain is both computationally efficient and easily parallelizable.
Once the Fourier transforms have been computed the now substantially compressed frequency domain representation can be collected on a single machine for further analysis.
We then proceed with the elimination of fractional poles by a simple element-wise multiplication.
No supplementary communication is needed to erase LRD and therefore the size of the data transmitted across the network is just $O(\frac{1}{V} d)$ as opposed to $O\left(N d \right)$. 
This remarkable improvement in communication requires only a modest computational cost of $O(\frac{1}{V})$ projections per data point on slave machines.

The compute time therefore allows an interactive experience for the user and becomes even shorter with a distributed implementation on several machines.
For example, on a single processor 
with a 2013 MacbookPro Retina we were able to compute $3000$ projections on $10^5$ samples in roughly a minute.

\begin{tabular}{c | c | c}
    Method & Time on slace & Comm. size \\
    \hline
    Time domain & $O(N d^2)$ & $O(N d + d^2)$\\
    \hline
    Fourier projection & $O(N d \frac{1}{V})$ & $O(d \frac{1}{V})$
\end{tabular}

\par{\textbf{{Memory:}}
A potential concern with the frequency domain approach is that the aggregation of the Fourier projections to a single device could exceed the device's memory.
The device will have to store projections of size $O(\frac{1}{V} d)$, compute element-wise products with time complexity $O(d^2 \frac{1}{V})$, and store the cross-correlation estimates in a memory container of size $O(d^2)$.
In particular, the maximum size of the memory needed by the algorithm on the master is $O(\frac{1}{V} d^2)$ which is small relative to the size of the data set in our current setting. Indeed, we assume that $N$ is large enough and therefore $\frac{1}{V} << N$.

\vspace{-0.2cm}

\section{Causality estimation on actual data}

\vspace{-0.2cm}
Identifying leading components on the stock market is insightful in terms of assessing which stocks move the market and highlights the characteristic latency of trading reactions. Consider two stocks, for instance AAPL and IBM (shares of Apple Inc. and IBM traded in the New York stock exchange). The trade and quote table of Thomson Reuters records all bids, asks, trades in volume and price. It is therefore interesting to check if there is a causation link between the price at which AAPL is traded as compared to that of IBM. In particular, if we see an increase in the price of the former can we expect an increase shortly after in the later? With which delay? Weak causation or short delay indicates an efficient market with few arbitrage opportunities. Significant causation and longer delays would enable high frequency actors to take advantage of the causal empirical relationship in order to conduct statistical arbitrage \cite{abergel2012market}.

\subsection{Causal pairs of stocks}
One critical application of the generic method we present is identifying which characteristic delays the NYSE stock market features as well as Lead-Lag ratios between pairs of stocks.
Lead-Lag ratios that are significantly different from $1.0$ indicate that changes in the price of one stock trigger changes in the price of another. 
This indicates pair arbitrageurs are most likely using high frequency arbitrage strategies on this pair of stocks.

\subsection{Using Full Tick data}
In order to highlight significant cross-correlation between pairs of stocks, one needs to consider high frequency dynamics. As we will show in the following, cross-correlation vanishes after a few milliseconds on most stocks and futures. 
In these settings it is then necessary to use full resolution data which in this instance comes in the form of Full Tick quote and trade tables (TAQ).
These TAQ tables record bids, asks and exchanges on the stock market as they happen. 
The timestamps are therefore irregular and not common to different pairs of stocks. 
Also, stock prices are Brownian motions and therefore feature long memory. 
This context is therefore in the very scope of data intensive tasks we consider. 
We show our novel Fourier compression based cross-correlation estimator provides consistent estimates in this setting.

\subsection{Checking the consistency of the estimator}
Consider ask and bid quotes during one month worth of data. We create a surrogate noisy lagged version of AAPL with a $13$ms delay and $91\%$ correlation which is named AAPL-LAG. We study fours pairs of time series: APPL/APPL-LAG, AAPL/IBM, AAPL/MSFT, MSFT/IBM. 
We study the changes in quoted prices (more exactly, volume averaged bid and ask prices). 
We obtained quote data for these stocks at millisecond time resolution representing several months of trading. 
We removed observations with redundant timestamps.
The cross-correlograms obtained below are computed between 10 AM and 2PM for $61$ days in January, February and March 2012. For each process, $3000$ frequencies were used in the Fourier basis
which is several orders-of-magnitude less than the number of observations that we get per day which ranges from $5 \times 10^4$ to $1 \times 10^5$.
The estimate cross-correlograms in Figure \ref{fig:emp_cross_corre} and their empirical significance intervals show that our estimator is consistent and does not suffer from non-vanishing variance as a result of LRD. 
We observe an $89\%$ average peak cross-correlation with an $8$ms delay for the surrogate pair of AAPL stocks which confirms our estimator is reliable with empirical data. 
While we observe the Fourier compression artifacts,
these only occur because our surrogate data features a deterministic delay. They do not affect pairs of actual observed processes.
In Figures \ref{fig:emp_cross_corre} and \ref{fig:emp_causation} we highlight a taxonomy of causal relationships and show in particular that with our definition of causality anchored in linear predictions, a process may cause another one without any significant delay. This may also be symptomatic of a delay shorter than the millisecond resolution of our timestamps.

\begin{figure}[h!]
\center
\begin{tabular}{cc}
    \includegraphics[width = 4cm]{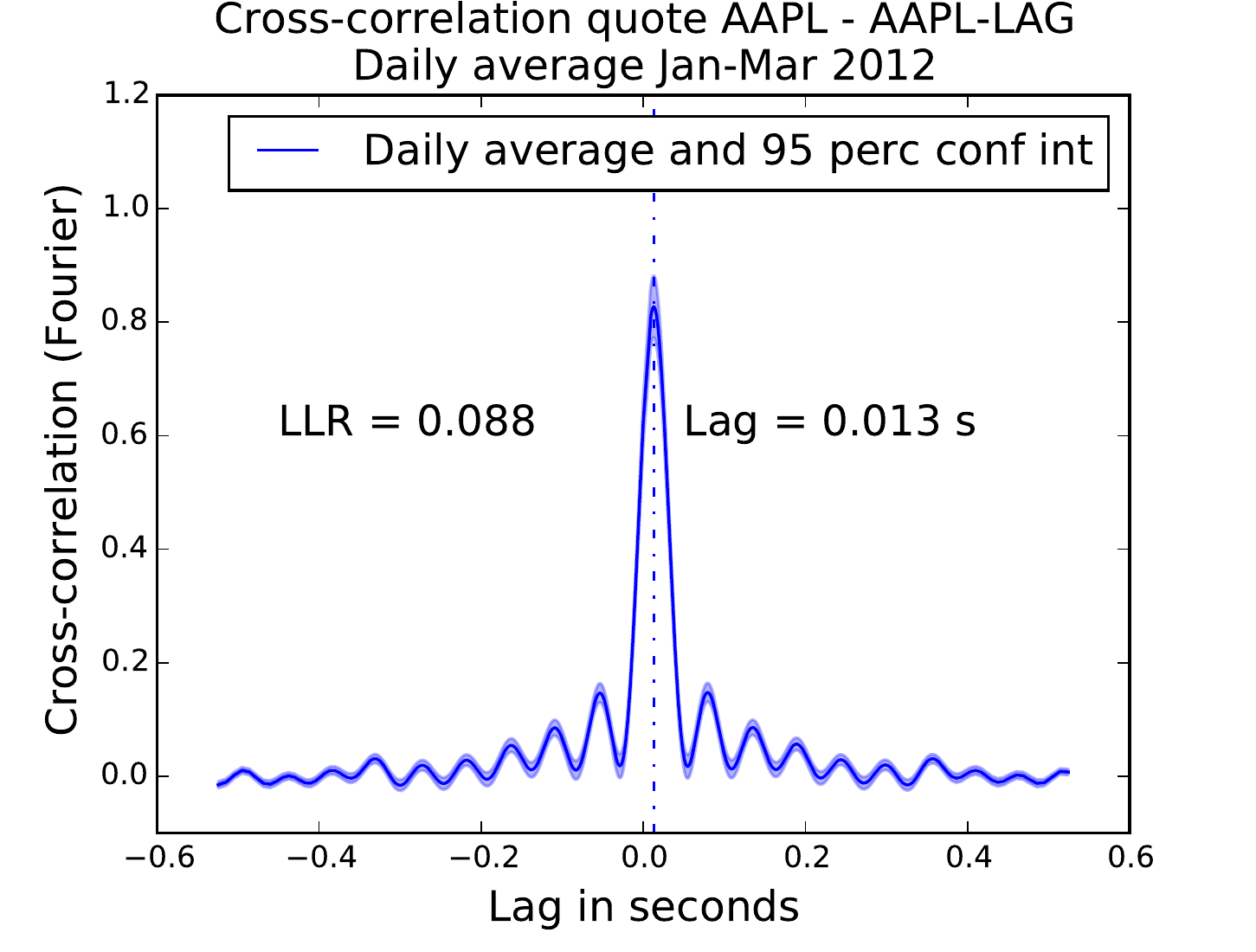} &  
    \includegraphics[width = 4cm]{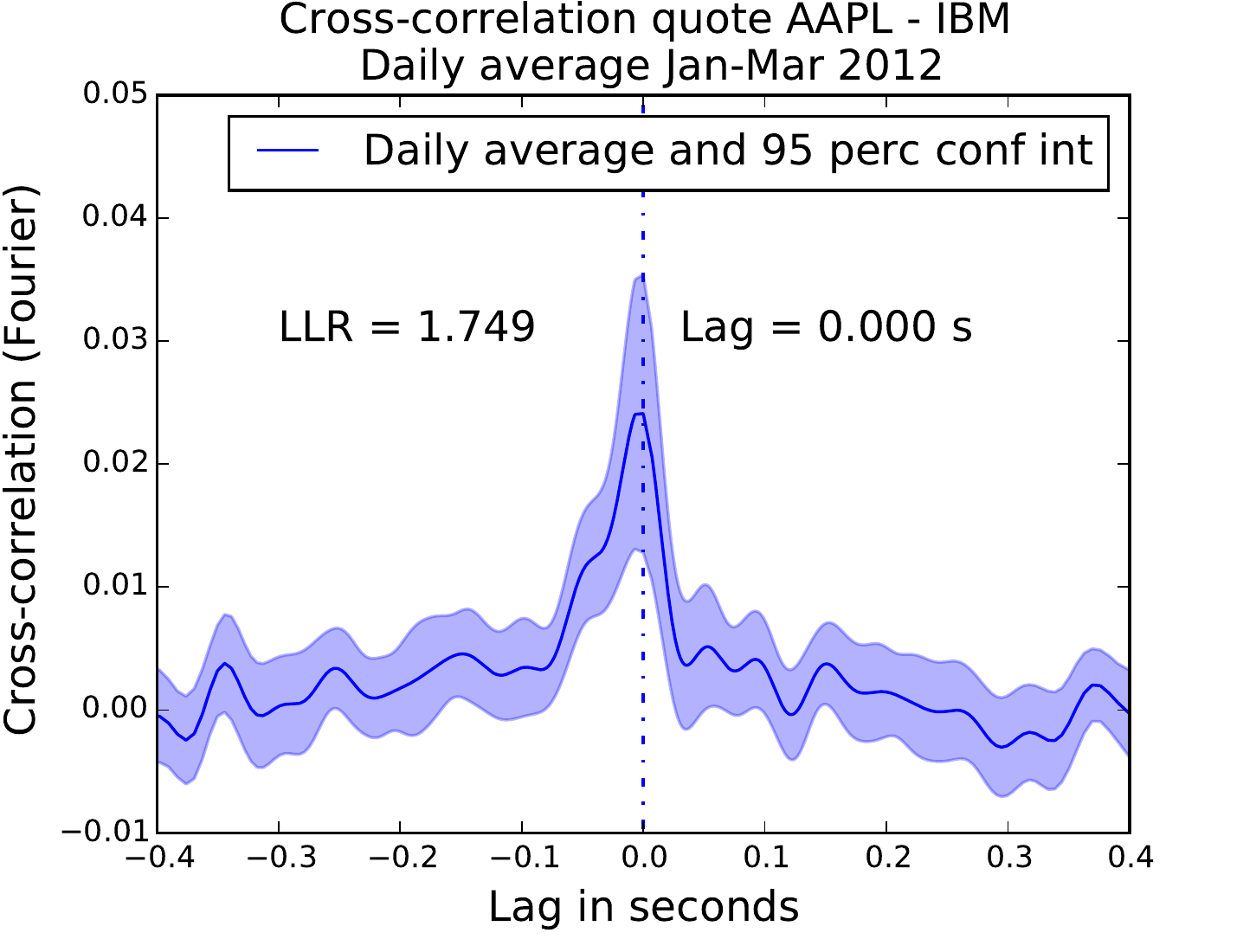} \\
    \includegraphics[width = 4cm]{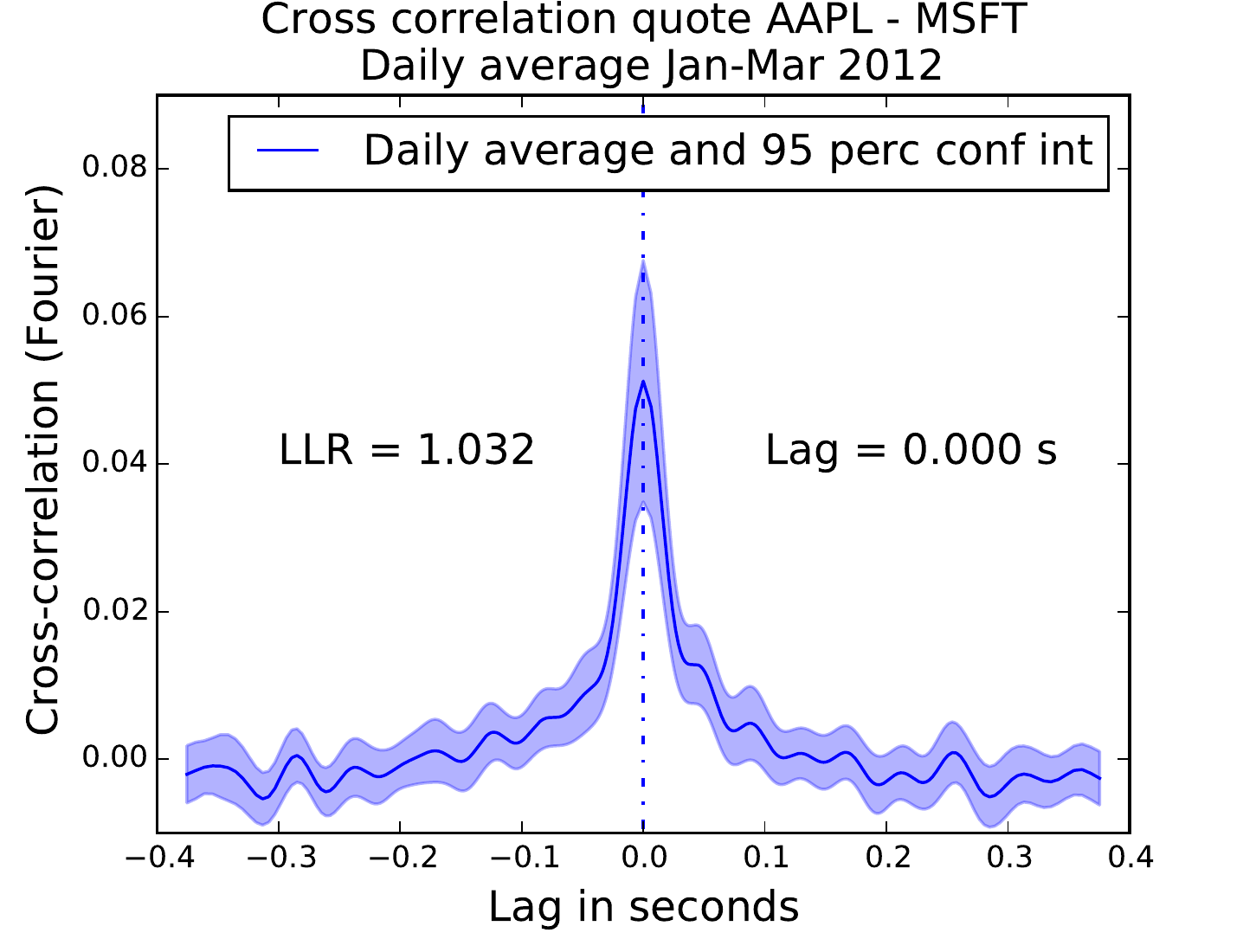} &  
    \includegraphics[width = 4cm]{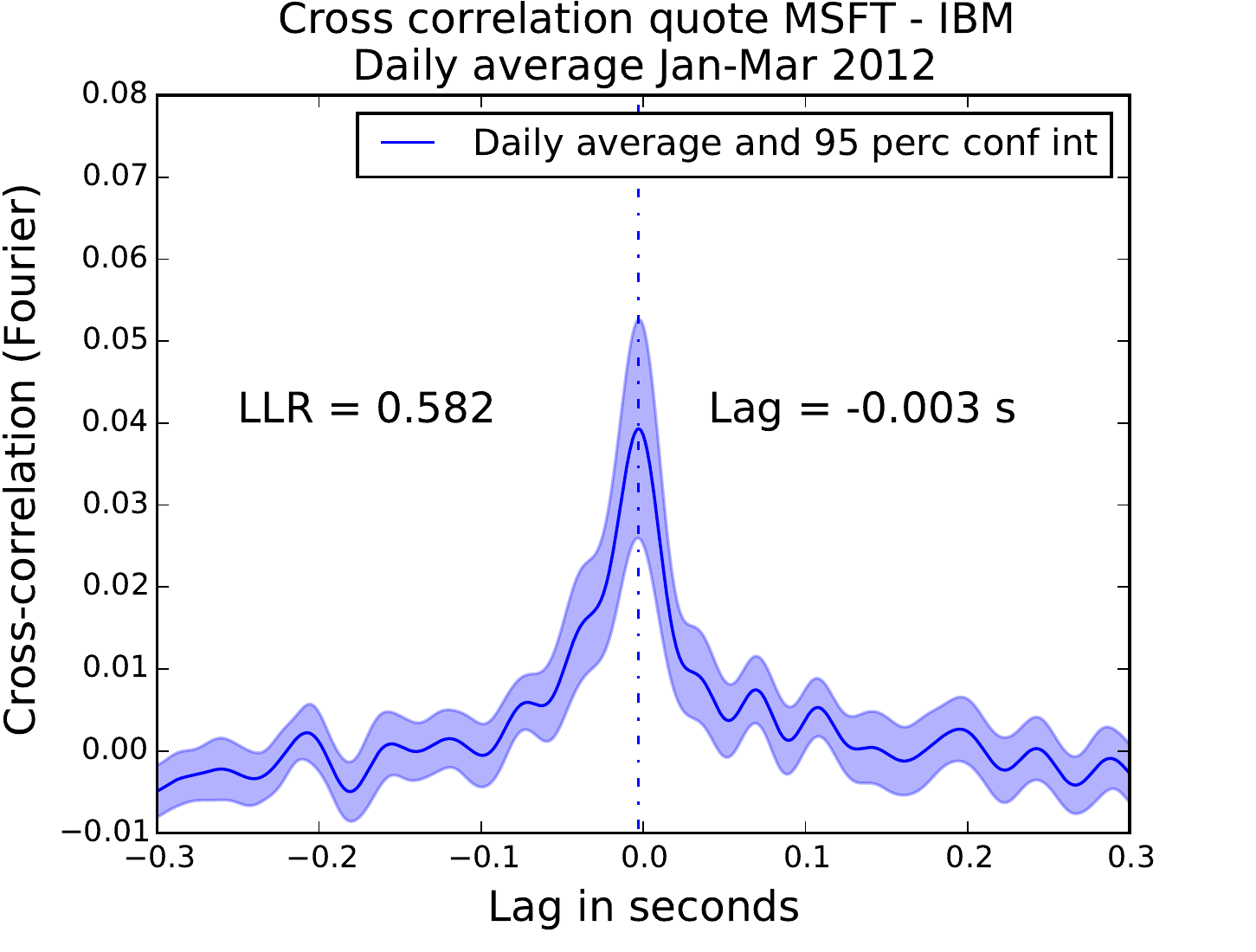}
\end{tabular}
\caption{
\footnotesize{
Average of daily cross-correlograms pairs of stock trade and quote data. Compression ratio is $<5\%$. We retrieve lag and correlation accurately on surrogate data. The daily averaged cross-correlogram of AAPL and IBM is strongly asymmetric, therefore highlight that AAPL causes IBM. The symmetry between AAPL and MSFT shows there is no such relationship between them. Finally, symmetric and offset in correlation peak show that MSFT causes IBM with a millisecond latency.}
}
\label{fig:emp_cross_corre}
\vspace{-0.3cm}
\end{figure}

\subsection{Choosing the number of projections}
In order to guide practitioners in their choice of the number of Fourier basis elements to project onto, we conduct a numerical experiment on actual data. We compute an empirical standard deviation of the daily cross-correlogram obtained in January 2012 (19 days) for JPM (JP Morgan Chase) and GS (Goldman Sachs) with $10$, $100$, $1000$ and $10000$ projections. Figures \ref{fig:emp_causation} and \ref{fig:Scalability} show that, as expected, the variance decreases linearly with the number of projections and we can obtain reliable estimates with $1000$ projections.

\begin{figure}[t!]
\center
\begin{tabular}{cc}
    \includegraphics[width = 4cm]{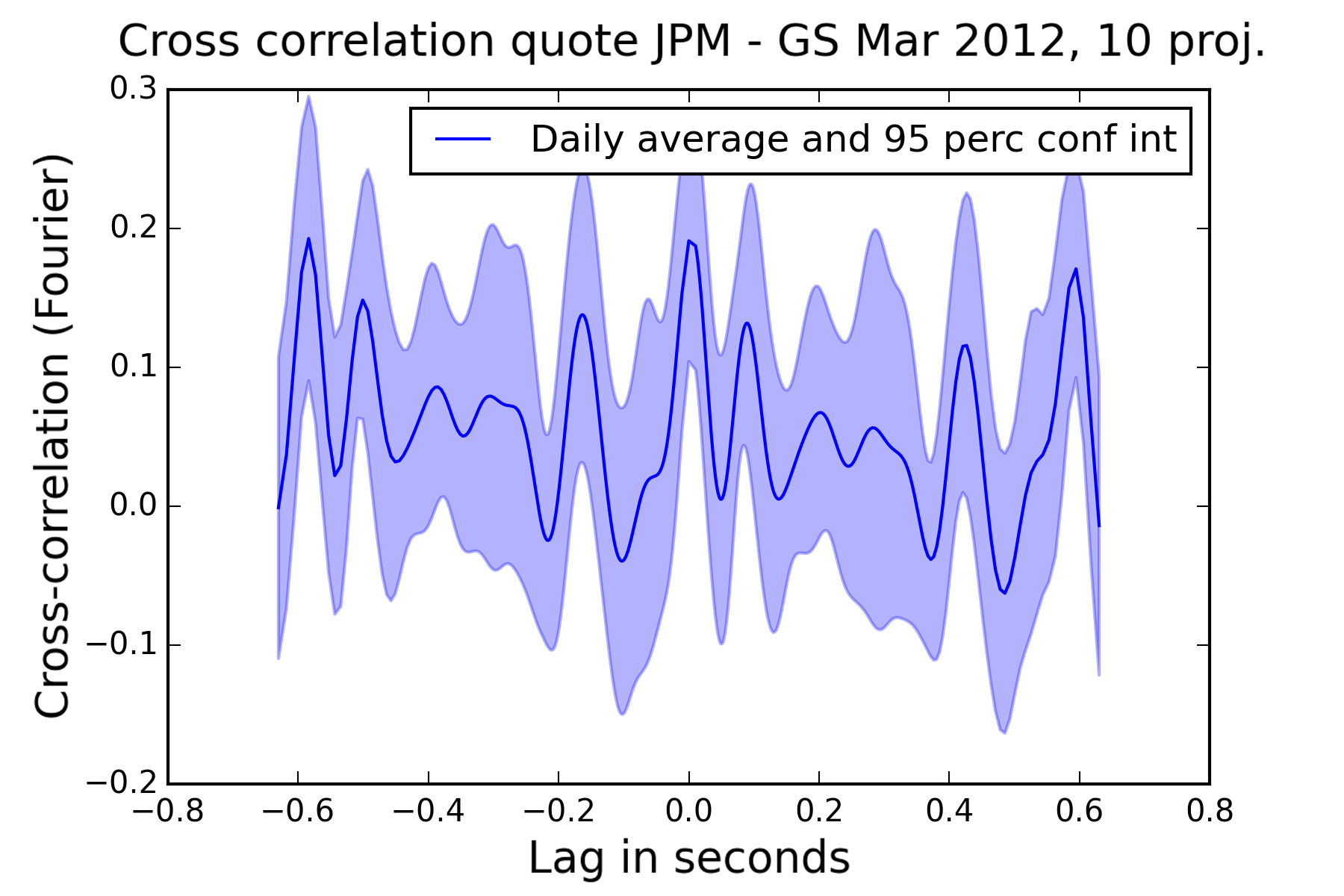} &  
    \includegraphics[width = 4cm]{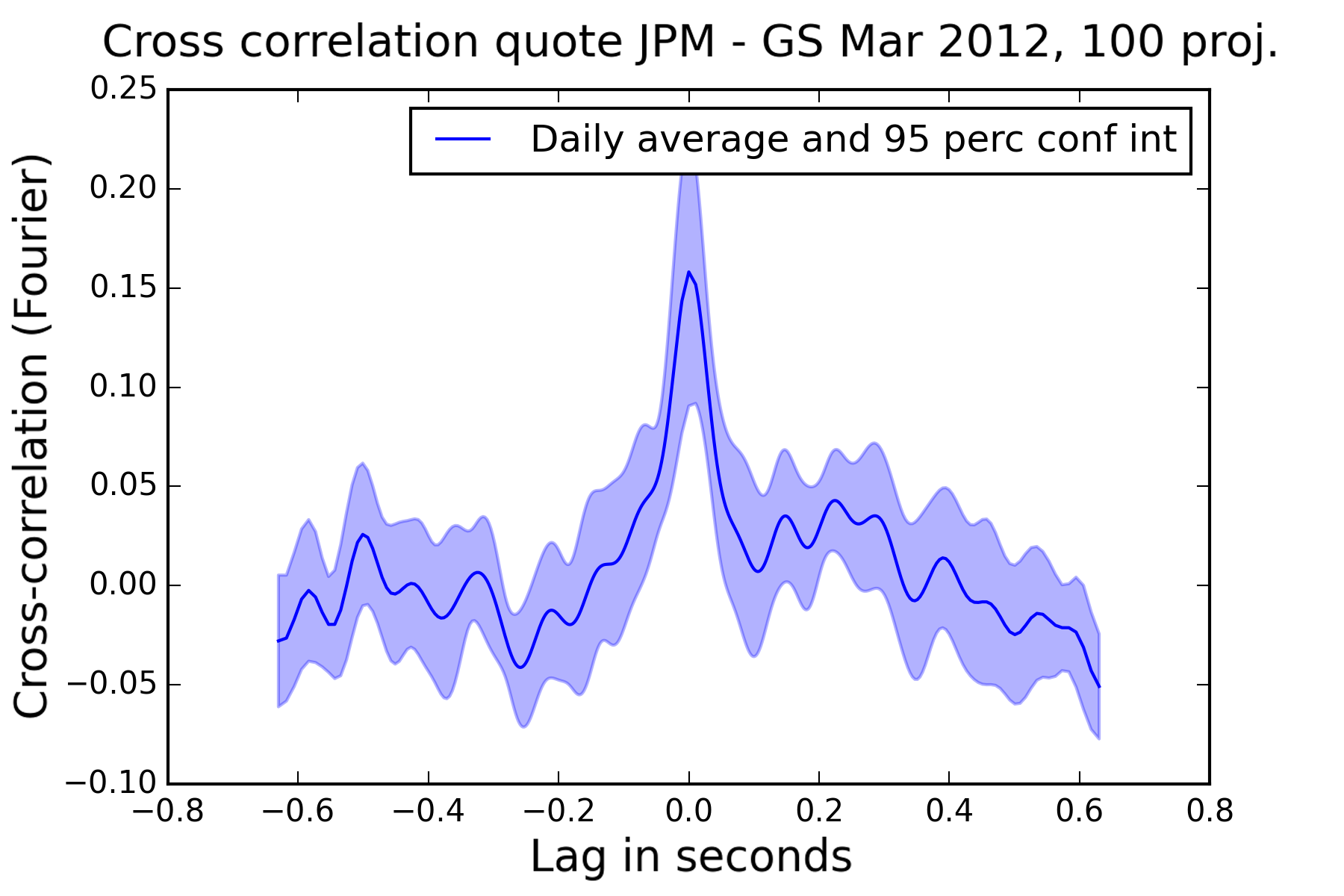} \\
    \includegraphics[width = 4cm]{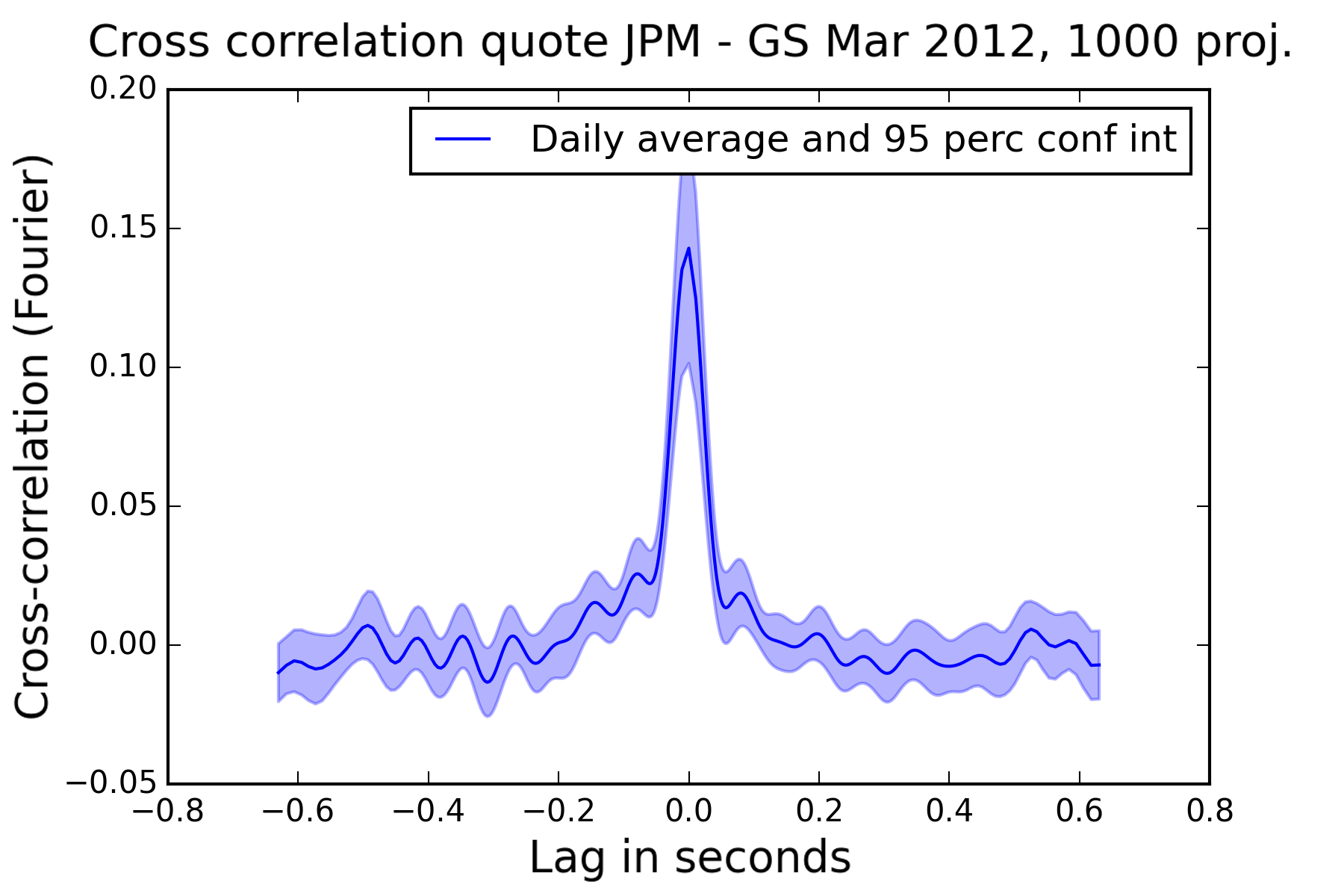} &  
    \includegraphics[width = 4cm]{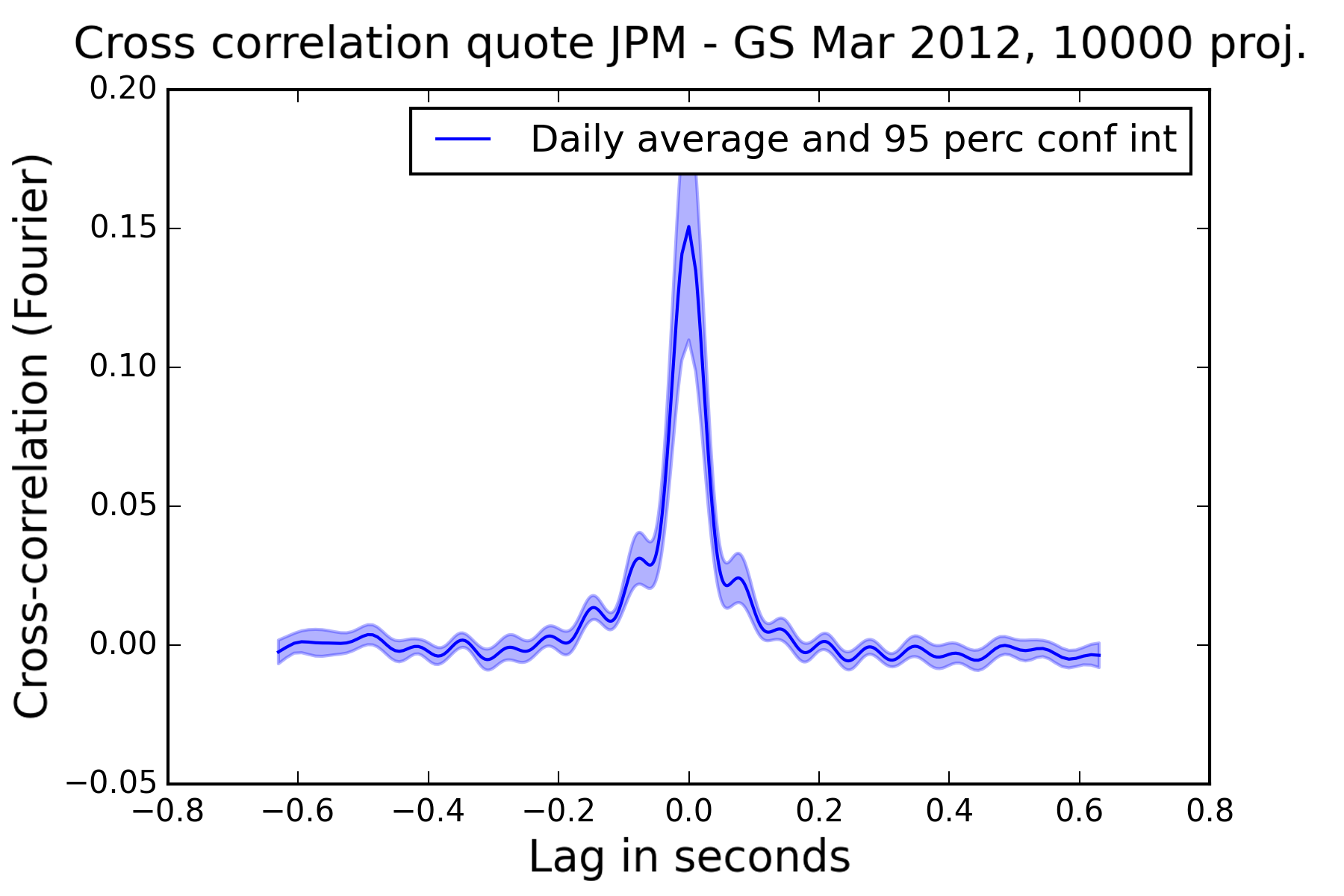}
\end{tabular}
\caption{
\footnotesize{
The empirical variance of cross-correlogram of JPM and GS computed daily over the month of January 2012 decreases with the number of projections we choose, a number between $10^3$ and $10^4$ is comfortable. This is to be compared with the $5 \times 10^4$ to $10^5$ samples taken into account each day, or $3 \times 10^6$ to $6 \times 10^6$ samples over the 60-trading day window of study. The standard deviation of each cross-correlogram is represented in Figure \ref{fig:Scalability}.
}
}
\vspace{-0.3cm}
\label{fig:emp_causation}
\end{figure}

\subsection{Studying causality at scale}
A primary goal of this work is to enable practical scalable causal inference for time series analysis. 
% The aim of the approach we present here is to enable analytics with large numbers of records. 
To evaluate scalability in a real-world setting in which $\frac{1}{V} << N$, we assess the relation between AAPL and MSFT over the course of $3$ months. 
In contrast to our earlier experiments (shown in Figure~\ref{fig:emp_cross_corre}), we no longer average daily cross-correlograms in and therefore only leverage concentration in the inverse Fourier transform step of the procedure.
With only $3000$ projections for $~5 \times 10^6$ observations per time series, the results we obtain on Figure~\ref{fig:Big_exp} reveals the causal relation between AAPL, AAPL-LAG, IBM and MSFT consistently with Figure~\ref{fig:emp_cross_corre}.

\begin{figure}[h!]
\center
\begin{tabular}{cc}
    \includegraphics[width = 3.8cm]{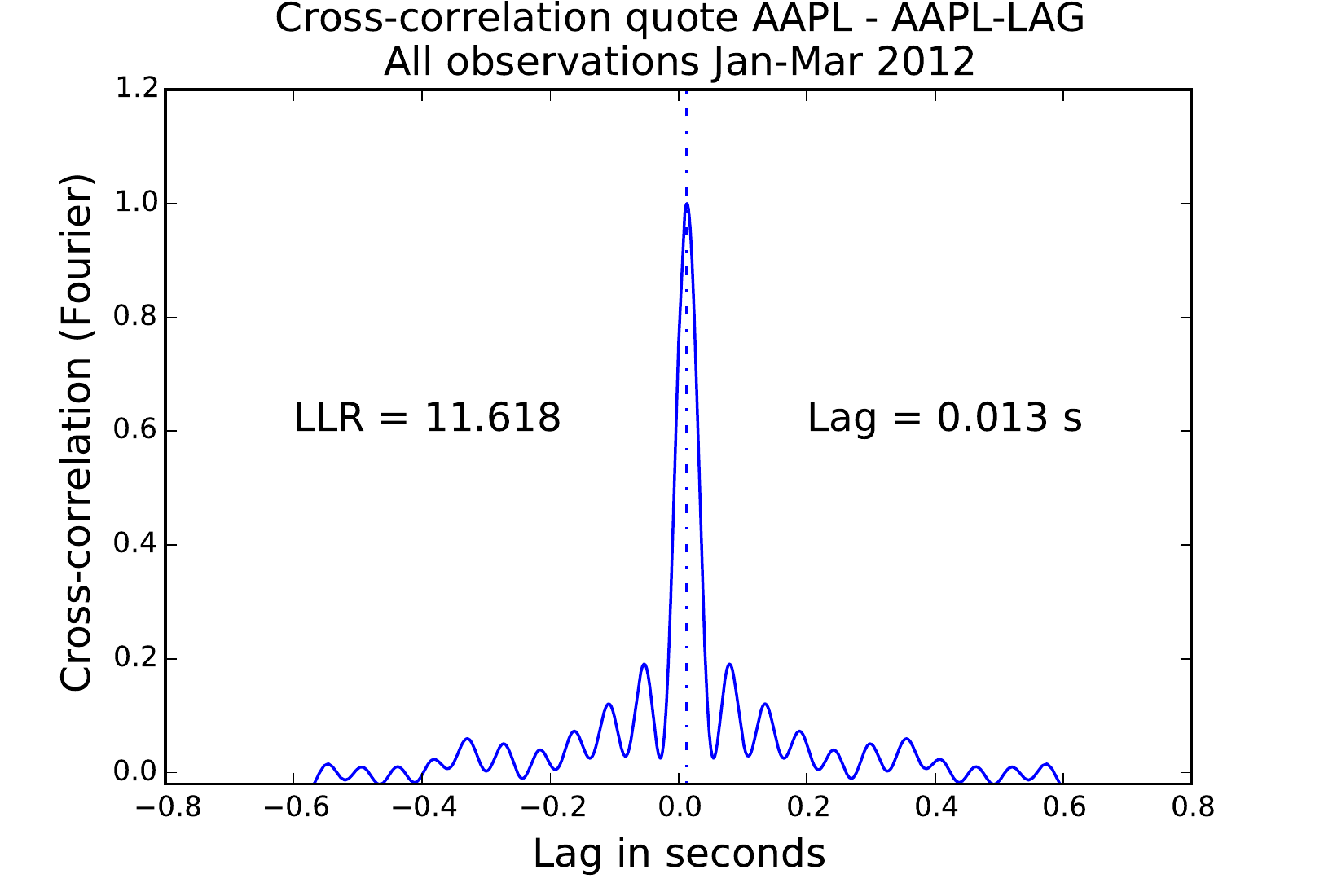} &  
    \includegraphics[width = 3.8cm]{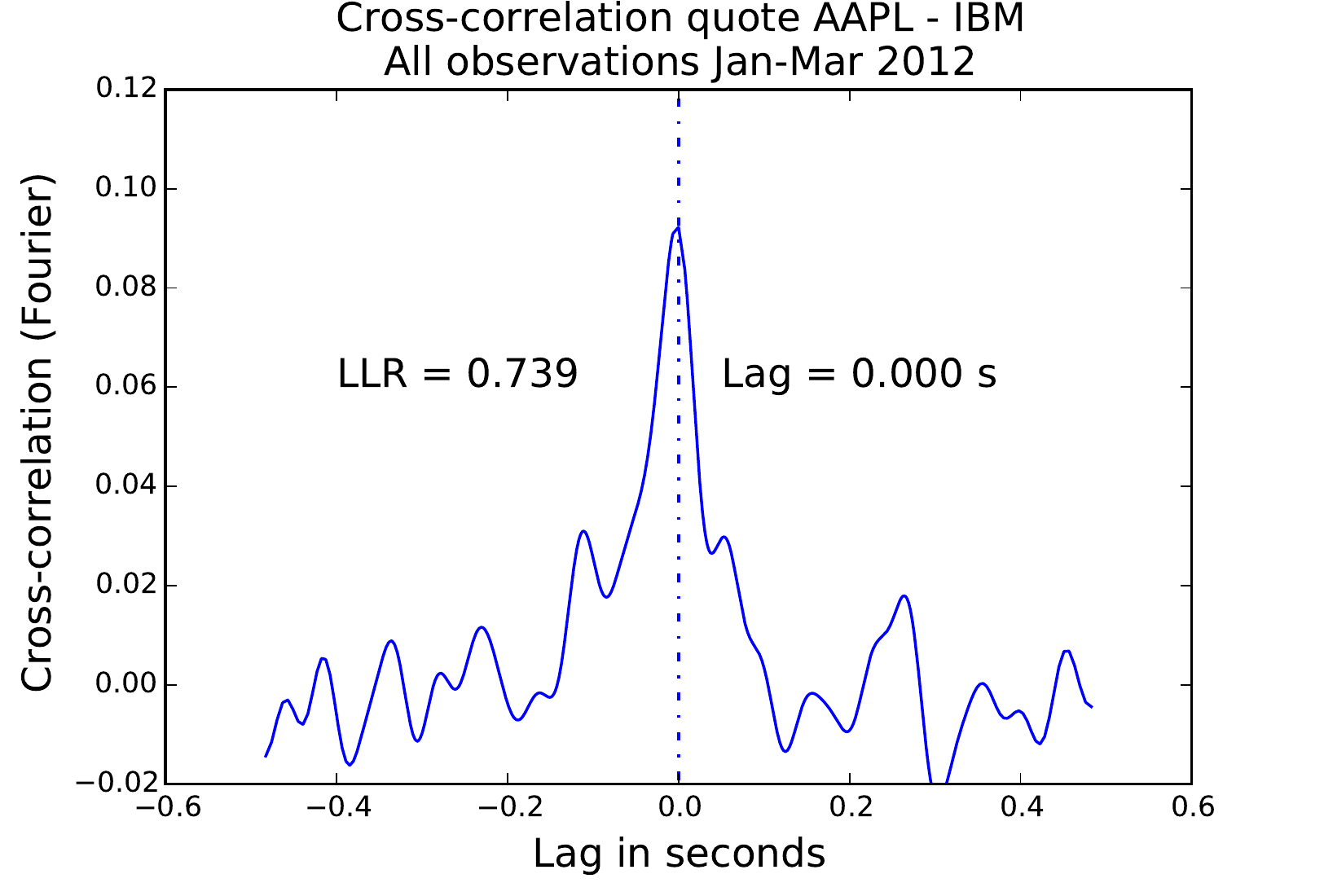} \\
    \includegraphics[width = 3.8cm]{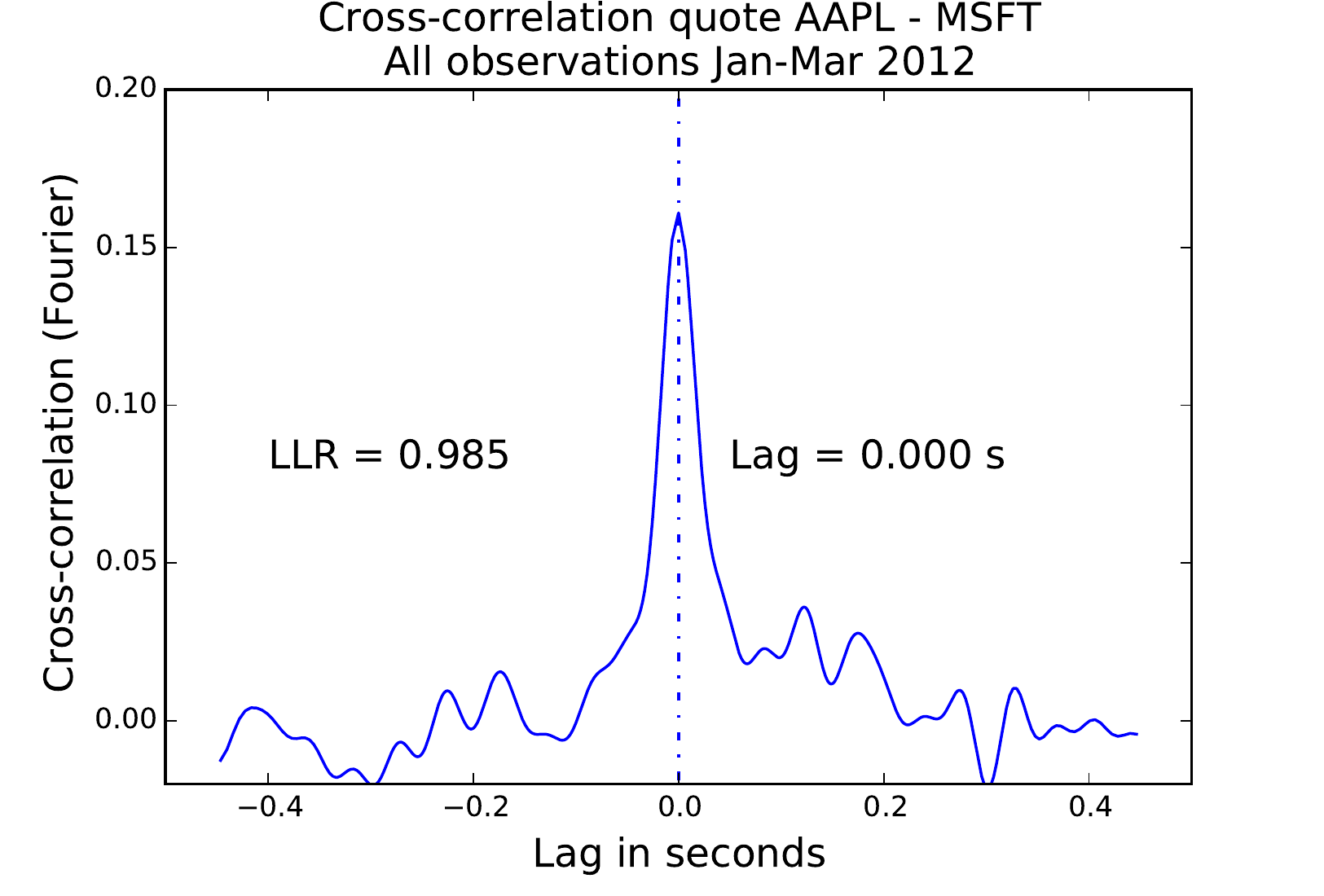} &  
    \includegraphics[width = 3.8cm]{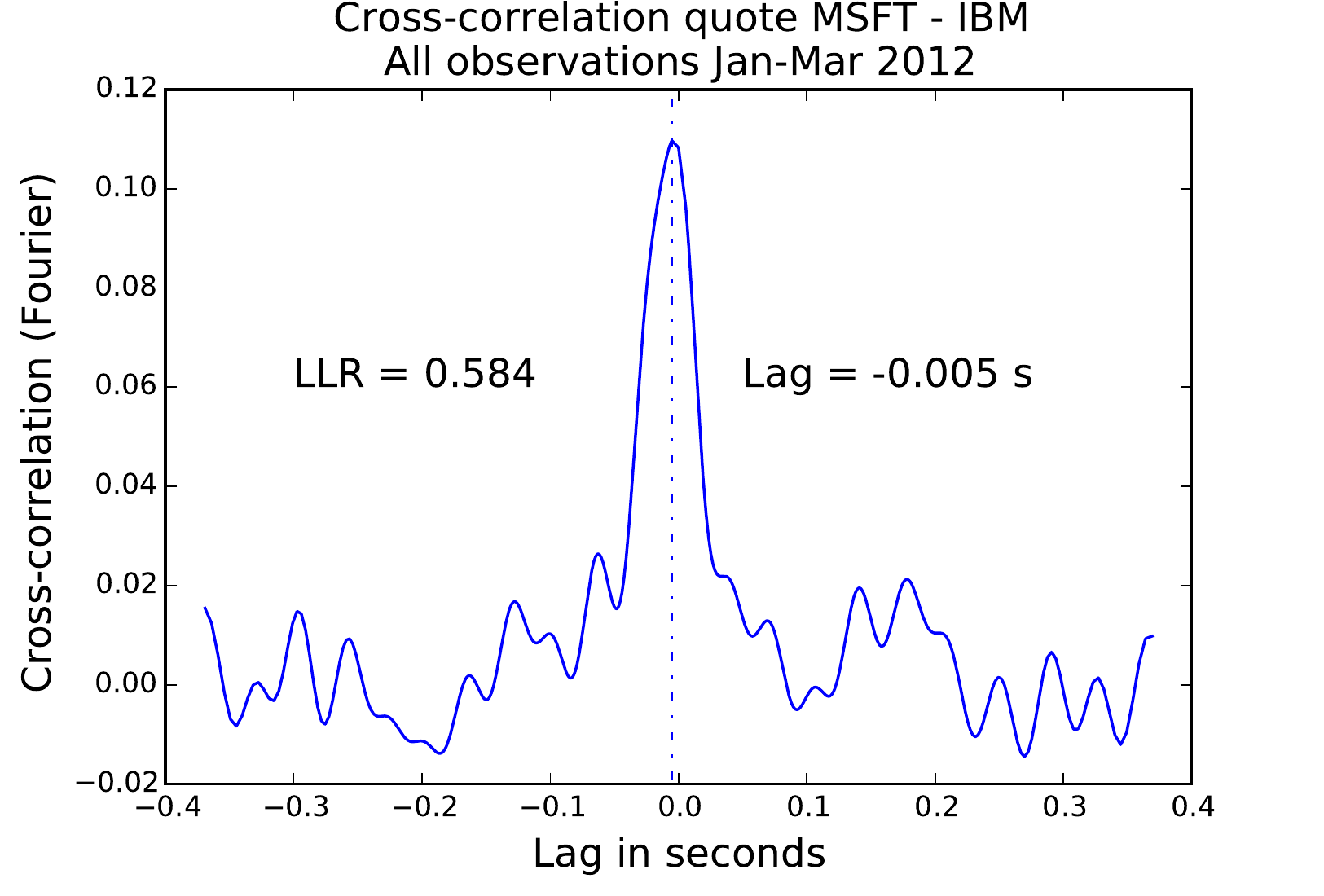}
\end{tabular}
\caption{
\footnotesize{
Compression ratio is $<1\%$. On the entire data set we retrieve results similar to \ref{fig:emp_cross_corre} therefore validating the use of our estimation of cross-correlograms in a scalable manner thanks to Fourier domain compression.}
}
\vspace{-0.3cm}
\label{fig:Big_exp}
\end{figure}

\par{\textbf{Scalability:}}
In order to assess the scalability of the algorithm in a situation where communication is a major bottleneck, we run the experiment with Apache Spark on Amazon Web Services EC2 machines of type r3.2xlarge. In Figure \ref{fig:Scalability} we show that even with a large number of projections ($10000$) the communication burden is still low enough to achieve speed-up proportional to the number of machines used.

\begin{figure}[h!]
    \centering
    \begin{tabular}{cc}
        \includegraphics[width = 3.8cm]{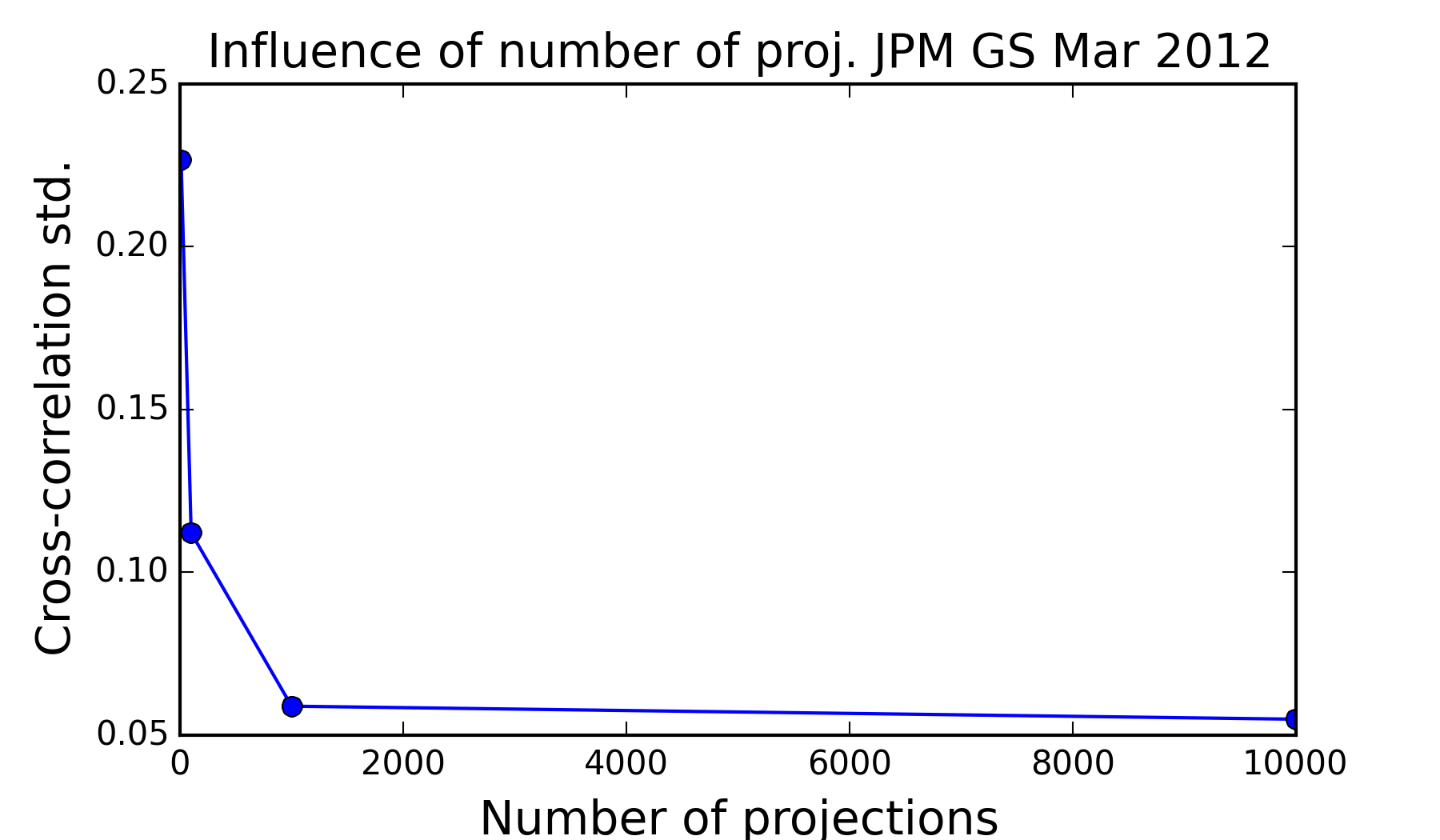} & \includegraphics[width = 3.8cm]{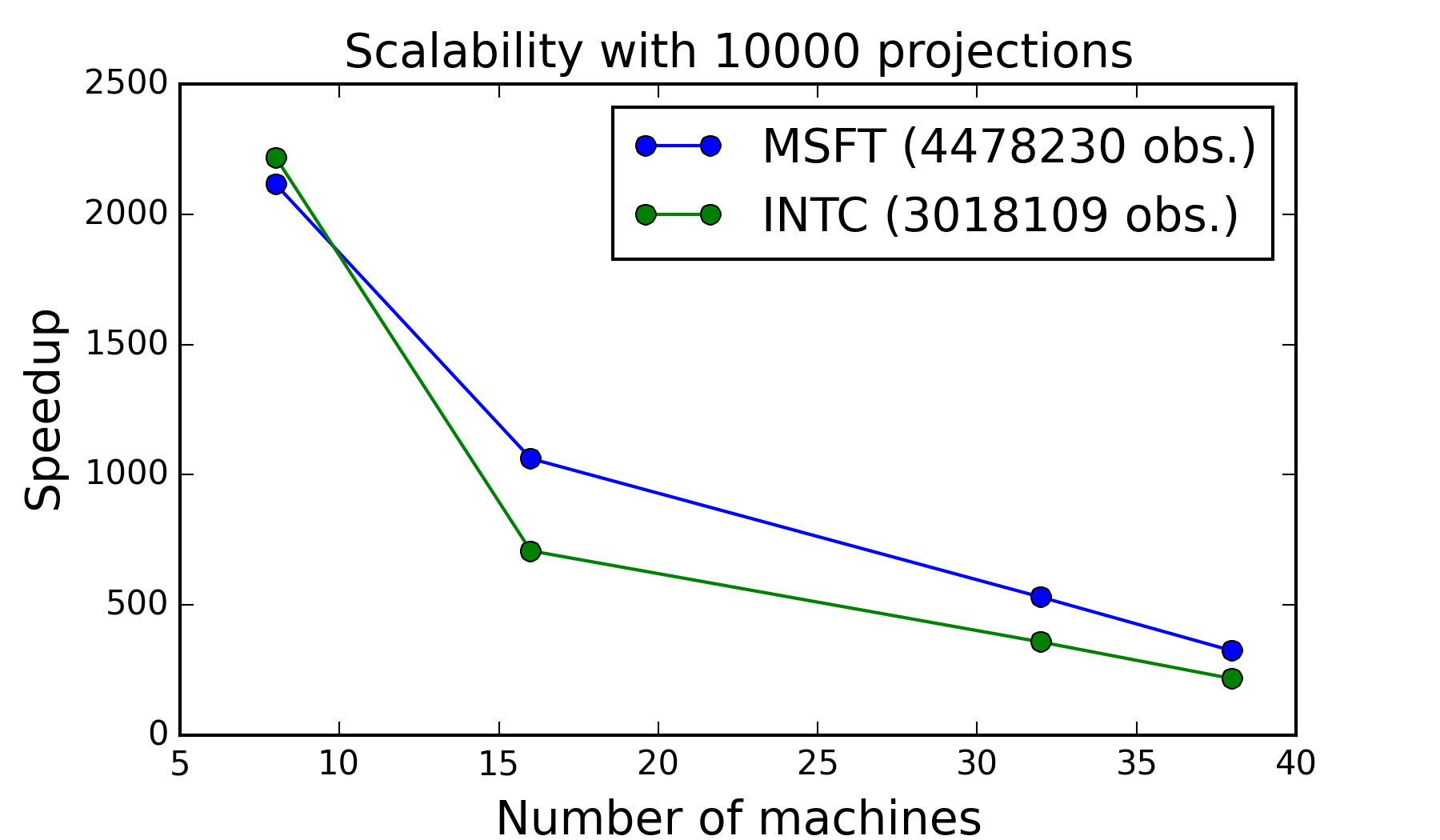}
    \end{tabular}
    \caption{
    \footnotesize{
    On the left we plot the empirical standard deviation of daily cross-correlograms (Figure \ref{fig:emp_causation}) with respect to the number of projections showing that the variability decreases rapidly. % after roughly 1000 projects.
    On the right we plot the run time performance of our algorithm % on $10^7$ records
    %with millions of records over January 2015 is plotted 
    versus the number of Apache Spark EC2 machines demonstrating approximately linear speedup.
    The small number of projections ($10^4$) relative to the size of the data set ($10^7$ records) avoids communication.
    }
    }
    \label{fig:Scalability}
\end{figure}

\vspace{-0.4cm}

\section{Conclusion}
Time series analysis via the frequency domain presents several presents unique opportunities in terms of providing consistent causal estimates and scaling on distributed systems.
We proposed a communication avoiding method to analyze causality which does not require any sorting or joining of data, works naturally with irregular timestamps without creating spurious causal estimates and makes the erasure of Long-Range dependencies embarrassingly parallel. 
Our approach is based on Fourier transforms as compression operators that do not modify the second order properties of stochastic processes. 
Applying an inverse Fourier transform to the resulting estimated spectra enables exploration of dependencies in the time domain. 
With the resulting consistent cross-correlogram, one can compute Lead-Lag ratios and characteristic delays between processes thereby infer linear causal structure. 
We show that projecting onto $3000$ Fourier basis elements is sufficient to study stock market pair causality with tens of millions of high frequency recordings, thereby providing insightful analytics in a generic and scalable manner.

\newpage

\par{\textbf{Acknowledgments:}}
The authors would like to thank Prof. David R. Brillinger for his advice on the theoretical aspects of the method we presented. This material is based upon work supported by the National Aeronautics and Space Administration under Prime Contract Number NAS2-03144 awarded to the University of California, Santa Cruz, University Affiliated Research Center.
This research is supported in part by NSF CISE Expeditions Award CCF-1139158, DOE Award SN10040 DE-SC0012463, and DARPA XData Award FA8750-12-2-0331, and gifts from Amazon Web Services, Google, IBM, SAP, The Thomas and Stacey Siebel Foundation, Adatao, Adobe, Apple Inc., Blue Goji, Bosch, Cisco, Cray, Cloudera, Ericsson, Facebook, Fujitsu, Guavus, HP, Huawei, Intel, Microsoft, Pivotal, Samsung, Schlumberger, Splunk, State Farm, Virdata and VMware.

\vspace{-0.2cm}

\bibliography{biblio.bib}
\bibliographystyle{IEEEtran}

\end{document}